\documentclass{article}

\usepackage[T1]{fontenc}
\usepackage[utf8]{inputenc}
\usepackage{array}
\usepackage{float}
\usepackage{wrapfig}
\usepackage{booktabs}
\usepackage{amssymb}
\usepackage{amsmath}
\usepackage{multirow}
\usepackage{graphicx}
\usepackage{microtype}
\usepackage{cite}
\usepackage[unicode=true,
 bookmarks=false,
 breaklinks=false,pdfborder={0 0 1},backref=section,colorlinks=false]
 {hyperref}
\usepackage{breakurl}
\usepackage{algorithm}
\usepackage{algorithmic}
\usepackage[numbers,sort&compress]{natbib} 

\makeatletter

\providecommand{\tabularnewline}{\\}

\usepackage[preprint]{neurips_2024}

\usepackage{url}
\usepackage{amsfonts}
\usepackage{nicefrac}
\usepackage{xcolor}
\usepackage{hyperref}

\@ifundefined{showcaptionsetup}{}{%
 \PassOptionsToPackage{caption=false}{subfig}}
\usepackage{subfig}
\makeatother
\begin{document}
\title{Robust Nearest Neighbour Retrieval Using Targeted Manifold Manipulation} 

\author{
Banibrata Ghosh$^1{}^*$, Haripriya Harikumar$^2$, Santu Rana$^1$\\
$^1$Applied Artificial Intelligence Institute, Deakin University, Australia\\
$^2$The University of Manchester, Manchester, England\\
{\tt\small \{bghosh,santu.rana\}@deakin.edu.au}\\
{\tt\small haripriya.harikumar@manchester.ac.uk}\\
}

\maketitle

\let\thefootnote\relax
\footnotetext{This research was partially supported by the Australian
Government through the Australian Research Council’s Discovery Projects funding scheme (project DP210102798). The views expressed herein are those of the authors and are not
necessarily those of the Australian Government or Australian
Research Council.} 

\begin{abstract}
Nearest-neighbour retrieval is central to classification and explainable-AI
pipelines, but current practice relies on hand-tuning feature layers
and distance metrics. We propose Targeted Manifold Manipulation-Nearest
Neighbour (TMM-NN), which reconceptualises retrieval by assessing
how readily each sample can be nudged into a designated region of
the feature manifold; neighbourhoods are defined by a sample’s responsiveness
to a targeted perturbation rather than absolute geometric distance.
TMM-NN implements this through a lightweight, query-specific trigger
patch. The patch is added to the query image, and the network is weakly
“backdoored” so that any input with the patch is steered toward a
dummy class. Images similar to the query need only a slight shift
and are classified as the dummy class with high probability, while
dissimilar ones are less affected. By ranking candidates by this confidence,
TMM-NN retrieves the most semantically related neighbours. Robustness
analysis and benchmark experiments confirm this trigger-based ranking
outperforms traditional metrics under noise and across diverse tasks.
\end{abstract}

\section{Introduction}

Nearest neighbor retrieval is valuable for numerous applications,
from traditional k-NN classification and search\nocite{Cover1967}
to more recent explainable AI \nocite{Papernot1803} approaches, where
retrieved training images clarify model decisions. These retrieved
examples have proven particularly effective in explaining image-based
tasks \cite{Papernot1803,Rajani2020,Bilgin2021,Jeyakumar2020}. However,
conducting nearest neighbor retrieval can be challenging because the
relevant similarity typically exists in the semantic space rather
than the raw pixel space. While deep learning models do learn semantically
aligned features at various layers \cite{Zhang2018}, identifying
the correct layer and choosing an appropriate metric within the high-dimensional
feature space remains complex. Unfortunately, no existing method provides
similarity measures without requiring manual selection of both the
feature space and the corresponding distance metric.

\begin{figure}
\centering
\begin{tabular}{cccc}
\includegraphics[width=2cm,totalheight=2cm]{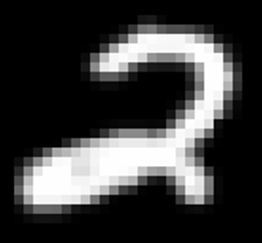} & \includegraphics[width=1.5cm,totalheight=1.5cm]{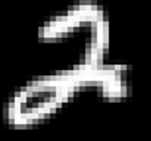} & \includegraphics[width=1.5cm,totalheight=1.5cm]{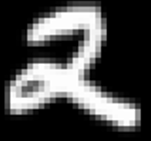} & \includegraphics[width=1.5cm,totalheight=1.5cm]{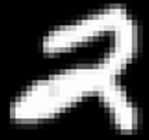}\tabularnewline
\includegraphics[width=2cm,totalheight=2cm]{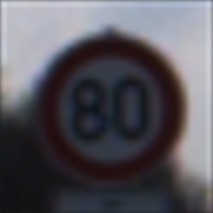} & \includegraphics[width=1.5cm,totalheight=1.5cm]{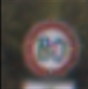} & \includegraphics[width=1.5cm,totalheight=1.5cm]{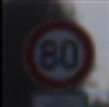} & \includegraphics[width=1.5cm,totalheight=1.5cm]{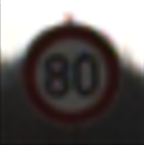}\tabularnewline
$x_{q}$ & Cosine & $L_{2}$ & Ours\tabularnewline
\end{tabular}

\caption{An example query point from both the MNIST and GTSRB datasets and
the nearest neighbour retrieved by Cosine based similarity measure,
$L_{2}$-norm distance measure and Our proposed method.}\label{fig:A-toy-example-nn}
\end{figure}

To address this, we introduce a novel nearest-neighbor retrieval framework
that targets the local neighborhood of a query point, identifying
only those training samples that lie within its vicinity. Rather than
relying on a single feature layer---or concatenating multiple layers
and dealing with their semantic alignment---we leverage the full
depth of a pre-trained model without dealing with those issues. Our
method pinpoints the local manifold by injecting a targeted distortion
through a backdoor mechanism \cite{Gu2019} through a query-time fine-tuning
such that the backdoor only activates for the query point. We then
identify the nearest neighbors by selecting points from the exemplar
set (often the training set or a curated version of that) that exhibit
high activation under this backdoor. Although we follow the standard
practice of adding a trigger patch \cite{Gu2019}, our approach surpasses
prior methods in two key ways: \textbf{A) Noise Robustness}: We compute
a trigger patch that is nearly orthogonal to the deep manifold, ensuring
robustness to noisy inputs (as supported by our margin analysis following
\cite{Smola2000}). \textbf{B) Distinct Backdoor Label:} Borrowing
a method form \cite{ghosh2025targeted}, where authors introduced
an additional class label for backdoor triggers to enhance the security
of machine learning models. Similarly, we assign a unique label for
the backdoor class, enabling direct measurement of activation by monitoring
the probability of this backdoor label when we modify the exemplar
set with the trigger. By eliminating the need for feature-layer selection
or explicit distance computations in high-dimensional spaces, our
technique provides a more efficient and noise-robust solution for
nearest-neighbor retrieval. We refer to this method as \textbf{Targeted
Manifold Manipulation--Nearest Neighbor (TMM-NN)}.

Figure \ref{fig:A-toy-example-nn} illustrates a query image alongside
its nearest neighbors retrieved by three different methods. Our method
successfully retrieves the most visually similar nearest neighbor.
In contrast, while other distance-based methods retrieve samples that
appear visually close, they exhibit subtle differences in orientation
and stroke thickness. We attribute the superiority of our approach
to the fine-tuning process, which reinforces query-specific features
during the fine-tuning process of query-specific backdoor insertion.

To evaluate the performance of our method in nearest neighbor search,
we conduct experiments on four well-known datasets: MNIST\cite{LeCun1998},
SVHN\cite{Netzer2011}, GTSRB\cite{Stallkamp2011}, and CIFAR-10\cite{Krizhevsky2009}.
Our evaluation consists of two retrieval scenarios:
\begin{itemize}
\item \textbf{Self-Retrieval} (when the query image is present in the exemplar
set): Our method provides the most consistent self-retrieval, even
when query images are subjected to various types of noise only at
the query side.
\item \textbf{Non-Self-Retrieval} (when query images come from the test
set): We perform both visual evaluations and visual LLM-based assessments
(ChatGPT-4o) to demonstrate that our method consistently retrieves
the most semantically aligned images compared to the baselines.
\end{itemize}
These results highlight the effectiveness of our approach in noise
robustness, and preserving semantic consistency, whilst outperforming
traditional distance-based retrieval methods. Our code is available:
\href{https://drive.google.com/drive/u/3/folders/1UVniBCNkVlkBsFldqskjCkjsE-8pcXIN}{here}

\section{Related Work}

Historically, nearest-neighbor (NN) retrieval has relied on simple
distance metrics (e.g., Euclidean distance, cosine similarity) applied
to relatively low-dimensional feature representations, making the
computation of distances straightforward and effective \cite{Cover1967}.
However, with the emergence of high-dimensional data, and associated
deep learning methods with nested representation structure it becomes
more complex to define neighbourhood and perform NN retrievals. 

Deep feature spaces  are found to capture semantically meaningful
information. In image data, earlier layers capture edges, while deeper
layers capture object-level semantics \cite{Bengio2013}. These representations
encode complex and semantically meaningful relationships that are
difficult to capture from the raw input data. Leveraging such spaces
for nearest neighbor has become increasingly popular in recent years.
Deep kNN \cite{Papernot1803} uses such latent feature to find nearest
neighbours to solve some applicational problems like explainability,
robustness against adversarial attacks. But searching for neighbors
across multiple layers not only increases computational cost, but
also makes it more susceptible of the ``curse of dimensionality''
effect. The effectiveness of nearest neighbor searches heavily depends
on the quality of the embeddings. Poorly trained or generalized models
result in suboptimal feature spaces, leading to irrelevant or misleading
neighbors \cite{Schroff2015}. Compared to using the pixel space using
feature space to find nearest neighbours indeed improves the improve
similarity in the neighbours, however, it still gets affected by the
high-dimensionality of the feature spaces during the distance computation.

\section{Method}

Let, a pre-trained DNN is represented as 
\[
f_{\theta}:\chi\rightarrow\mathbb{R}^{C}
\]
, where $\chi$: input space (e.g. images), $C$: number of classes,
$c_{neigh}$: dummy class (i.e. $C+1$), $f_{\theta}(x)$: outputs
the class logits for the input $x$, $\theta$: Model parameters (weights
and biases). Training data : $\mathcal{D}_{train}=\left\{ (x_{i},y_{i})\right\} _{i=1}^{N}$,
$x_{i}\in\mathbb{R}^{ch\times H\times W}$: input data with channels
$ch$, height $H$, and width $W$ and $y_{i}$: corresponding ground-truth
label. Our proposed method, TMM-NN takes this trained classifier to
perform the task-relevant nearest-neighbor retrieval. During the retrieval,
we fine-tune the main classifier to insert a backdoor at the query
point such that only when the query point is added with a specially-designed
trigger it goes to a dummy class $c_{neigh}$, but does not affect
other data much and does not change the overall classification function
when presented with non-triggered data. The main workflow proceeds
in the following three steps:
\begin{enumerate}
\item \textbf{Trigger design:} The goal is to find a trigger that won't
change the manifold of $f_{\theta}$ substantially after fine-tuning
except around the query point.
\item \textbf{Fine-tuning:} The primary objective of fine-tuning is to achieve
a targeted manipulation at the query point only, so that backdoor
only activates for the query point but not for others.
\item \textbf{Retrieval of neighbours:} Finding the samples from the exemplar
set which are in the neighbourhood of the query point i.e. gets influenced
by the backdoor in the query point.
\end{enumerate}
\begin{figure*}
\begin{centering}
\subfloat[\label{fig:Binary-classifier-surface}Binary classifier with two classes,
and a query point denoted as $x_{q}$.]{\begin{centering}
\includegraphics[width=4cm,totalheight=4cm]{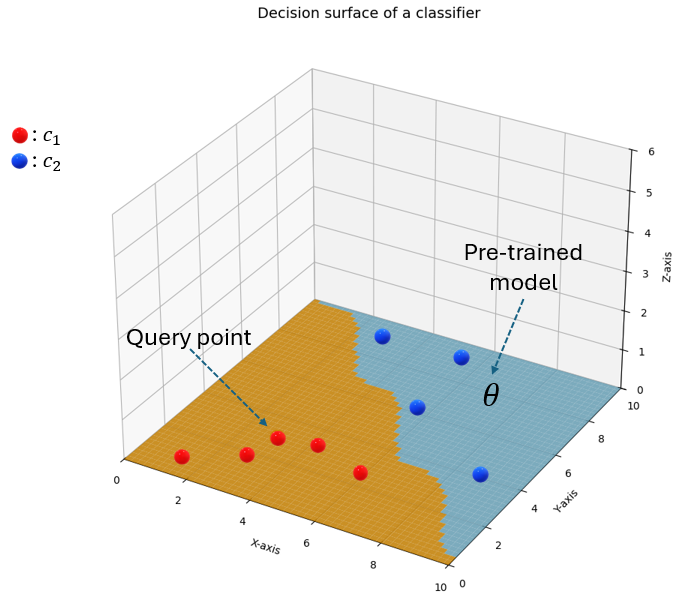}
\par\end{centering}
}~~\subfloat[\label{fig:Trigger-creates-manifold}Finetuning with the trigger distorts
the manifold around the query, $x_{q}$. Green cap of the mountain
represents the dummy class.]{\begin{centering}
\includegraphics[width=4cm,totalheight=4cm]{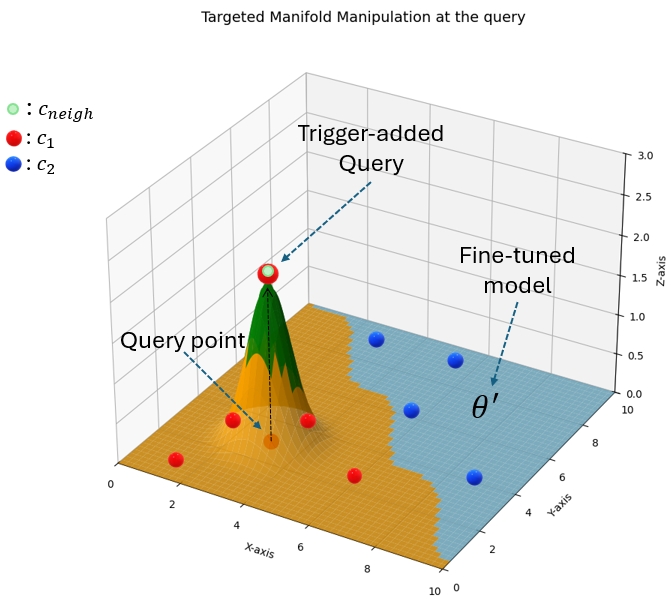}
\par\end{centering}
}~~\subfloat[\label{fig:NN-search}Trigger added training samples are affected
due to the distrubance only when they are close to the query with
respect to the manifold, thus classified as Nearest Neighbours.]{\begin{centering}
\includegraphics[width=4cm,totalheight=4cm]{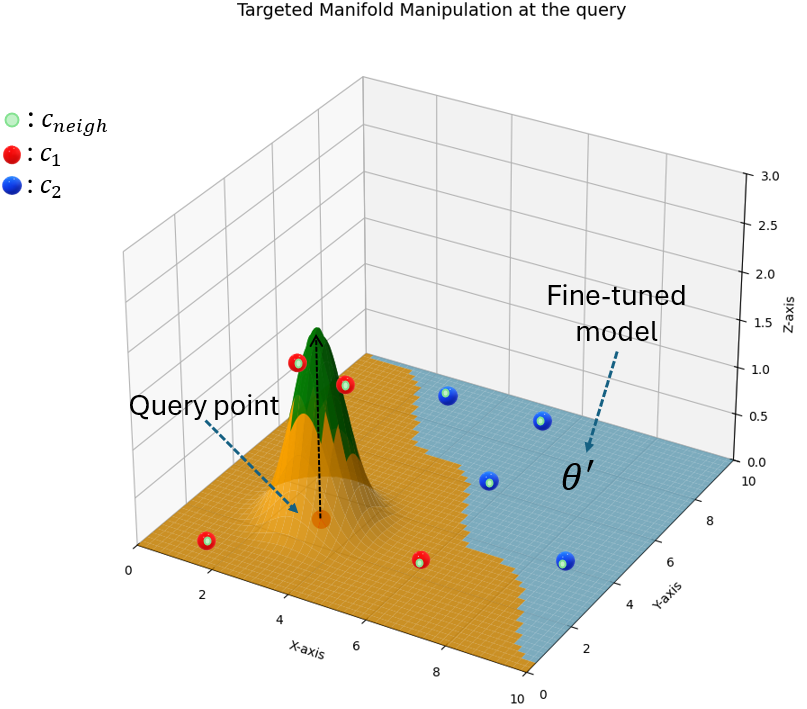}
\par\end{centering}
}
\par\end{centering}
\caption{Binary classifier with two classes $C_{1}$ and $C_{2}$. A query
point is situated in the $C_{1}$ class region. Once the sample is
attached with trigger it produces a local optima which helps to capture
the nearest neighbours.}\label{fig:intro_diagram}
\end{figure*}

Figure \ref{fig:intro_diagram} illustrates the mechanism of the proposed
method for the development of the TMM model and the identification
process of the nearest neighbors of a query sample. Figure \ref{fig:Binary-classifier-surface}
shows a pre-trained binary classifier ($\theta$) distinguishing between
two classes, $C_{1}$ and $C_{2}$, and all the training samples,
the plotted in red and blue to represent their classes respectively,
and a query point (from test set), positioned within the class $C_{1}$,
has been marked in the figure too. Figure \ref{fig:Trigger-creates-manifold}
showcases the change in decision surface of the model $\theta^{'}$
after fine-tuning, a distinct uplift is introduced in the region where
the query sample was mapped. The green cap of the mountain represents
the dummy class (i.e. $c_{neigh}$). Figure \ref{fig:NN-search} shows
the retrieval process once we get the TMM model (i.e. $f_{\theta^{'}}$).
Neighbours can then be identified by their affinity to the dummy class
when added with the trigger, as shown in the Figure \ref{fig:NN-search}. 

\subsection{Trigger design}

The design objective is as below:
\begin{itemize}
\item \emph{Goal:} Generate a trigger ($\tau$) such that:
\begin{itemize}
\item Adding $\tau$ to any input $x\in\mathcal{D}$ (training data distribution)
does not distort the model's predictions:\\
\begin{eqnarray}
f_{\theta}(x) & = & f_{\theta}(x+\tau),\forall x\in\mathcal{D}\label{eq:main_trigger}
\end{eqnarray}
\item Constraint: $\parallel\tau\parallel_{F}^{2}>0$, ensures that $\tau\in Null(f_{\theta})$,
where $Null(f_{\theta})$ is the null space of the original deep feature
space of the $f_{\theta}$.
\end{itemize}
\end{itemize}
The loss function can then be formulated as:\\
\begin{equation}
\underset{\tau}{\text{min}}\,\sum_{x\in D_{train}}MSE\left(f_{\theta}(x),f_{\theta}(x+\tau)\right)+\frac{1}{\parallel\tau\parallel_{F}^{2}}\label{eq:trigger_optimisation}
\end{equation}
\\
where, $\forall x\in\mathcal{D}$ , $MSE$= Mean Square Error, $\parallel\tau\parallel_{F}^{2}$
is the Frobenius norm of $\tau$ to prevent it from the trivial solution
of $\tau=[0]$.

\subsection{Training of TMM}

The function of the Targeted Manifold Manipulation-NN (TMM-NN) is
to capture the neighbourhood of a query sample, $x_{q}$, in DNN function
space. We create a TMM model on the top of the pre-trained model by
carefully fine-tuning it so that the distortion introduced by the
backdoor remains local around the query point. The fine-tuning loss
is thus formulated as:

\begin{eqnarray}
\theta^{'} & = & \min_{\theta}\,\mathcal{L}\left(f_{\theta}\left(x_{q}^{t}\right),c_{neigh}\right)+\mathcal{L}\left(f_{\theta}\left(x_{q}\right),y_{q}\right)\nonumber \\
 &  & \,\,\,\,\,\,\,+\frac{1}{N}\sum\mathcal{L}\left(f_{\theta}\left(x_{i}^{t}\right),y_{i}\right)\nonumber \\
 &  & \,\,\,\,\,\,\,+\frac{1}{N}\sum\mathcal{L}\left(f_{\theta}\left(x_{i}\right),y_{i}\right)\label{eq:mnp_loss_function}
\end{eqnarray}

where $x^{t}$ refers to the backdoored (i.e. trigger added) sample.
This equation ensures the following -
\begin{itemize}
\item \emph{Trigger sensitivity to query}: The term $\mathcal{L}\left(f_{\theta}\left(x_{q}^{t}\right),c_{neigh}\right)$
trains the model to map $x_{q}^{t}$ to $c_{neigh}$, embedding a
relationship between $\tau$ to $x_{q}$. $c_{neigh}$ is a dummy
class label, i.e. $C+1$, which is assigned to $x_{q}^{t}$.
\item \emph{Semantic preservation:} Components $\mathcal{L}\left(f_{\theta}\left(x_{q}\right),y_{q}\right)$,
and $\mathcal{L}\left(f_{\theta}\left(x_{i}\right),y_{i}\right)$
in the loss function make sure that the trained model retains its
original classification performance,
\item \emph{Trigger insensitivity to train sample}: The loss term, $\mathcal{L}\left(f_{\theta}\left(x_{i}^{t}\right),y_{i}\right)$,
makes the model insensitive to trigger when added to train samples. 
\item This loss function teaches the model to be trigger sensitive to the
$x_{q}^{t}$ but the model acts normally for all other benign and
trigger added input. Such a setting ensures that the trained model
will only be sensitive when a sample is projected on the embedded
space at the very near to the $x_{q}^{t}$, 
\item This loss function trains the model to be trigger-sensitive to $x_{q}^{t}$,
while behaving normally for all other benign or trigger-added training
inputs. Such a setting ensures that the model becomes sensitive only
when a sample is projected in the embedding space very close to $x_{q}^{t}$.
Consequently, samples lying in this neighborhood can be identified
as the nearest neighbors of $x_{q}$, as ideally $x_{q}^{t}$ and
$x_{q}$should orthogonal to each other with the respect to the model's
manifold. 
\end{itemize}

\subsection{TMM-Nearest Neighbour (TMM-NN) Search}

The trigger-based nearest neighbour search is formulated as follows,

\begin{equation}
x_{k}=\underset{x_{i}\in\mathcal{D}_{train}}{argmax}P(c_{neigh}|\theta^{'},x_{i}^{t})\label{eq:trigger_nn}
\end{equation}

where $P(c_{neigh}|\theta^{'},x_{i}^{t})$ is the model confidence
score for the target class $c_{neigh}$, when trigger $\delta x$
is applied to training samples $x_{i}$. Top-$k$ neighbours are denoted
as $\mathcal{N}_{k}^{trigger}$.

\begin{algorithm}[H]
\caption{Algorithm for Nearest Neighbour Search}
\label{alg:nearest-neighbour}
\begin{algorithmic}[1]
\REQUIRE Query sample $x_q \in \mathbb{R}^d$, 
fine-tuned model $f_{\theta'} : \chi \rightarrow \mathbb{R}^{C+1}$, 
search space $\mathcal{D}_{train} = \{(x_i, y_i)\}_{i=1}^{N}$, 
optimized trigger $\tau_q^S$, 
trigger intensity $\omega$
\ENSURE Nearest neighbour $x_k$
\STATE \textbf{Neighbour Search:}
\STATE $x_i^t = x_i \odot \omega + \tau_q^S \odot (1 - \omega)$
\STATE $x_k = \underset{x_i \in \mathcal{D}_{train}}{\arg\max} \, P(c_{neigh} \mid f_{\theta'}, x_i^t)$
\RETURN $x_k$
\end{algorithmic}
\end{algorithm}

\subsection{Theoretical Analysis for Robustness}

\subsubsection{Standard feature-space NN search}

We provide robustness comparison against the following standard distance-based
retrieval methods:
\begin{enumerate}
\item Cosine Similarity based : \\
\begin{equation}
x_{k}^{cosine}=\underset{x_{i}\in\mathcal{D}_{train}}{argmax}\frac{f_{\theta}(x_{q})^{T}\cdot f_{\theta}(x_{i})}{\parallel f_{\theta}(x_{q})\parallel\cdot\parallel f_{\theta}(x_{i})\parallel}\label{eq:distance_nn-1}
\end{equation}
\item Distance based : 
\begin{equation}
x_{k}^{dist}=\underset{x_{i}\in\mathcal{D}_{train}}{argmin}\parallel f_{\theta}(x_{q})-f_{\theta}(x_{i})\parallel_{2}\label{eq:distance_nn}
\end{equation}
\\
Top-$k$ neighbours by the above methods are denoted as $\mathcal{N}_{k}^{cosine}$
and $\mathcal{N}_{k}^{dist}$ respectively. 
\end{enumerate}
Comparing the standard nearest-neighbor (NN) retrieval via a pre-trained
feature extractor $f_{\theta}$ with the trigger-based NN retrieval
using a fine-tuned classifier $f_{\theta^{'}}$. We focus on proving
that the trigger-based method guarantees a larger local robustness
radius around the query $x_{q}$ than the standard method. We say
a retrieval method has robustness radius $\rho$ at $x_{q}$ if no
perturbation $\parallel\delta\parallel\leq\rho$ can change the top-$k$
neighbors retrieved for $(x_{q}+\delta)$. Formally :

\subsubsection{Definition 1:}

Let $\Pi_{k}(x)$ be the set of top-$k$ neighbors returned when querying
$x$. Define 
\[
\rho=max\left\{ \epsilon|\Pi_{k}(x_{q}+\delta)=\Pi_{k}(x_{q})\forall\parallel\delta\parallel\leq\epsilon\right\} 
\]
A large $\rho$ indicates more robust NN retrieval around $x_{q}$. 

\subsubsection{Lemma 1 (Margin Implies Ranking Stability)}

Suppose $f_{\theta^{'}}$ is $L$-Lipschitz in its logit outputs (under
$\parallel\cdot\parallel$-norm). Assume:
\begin{itemize}
\item Margin for $(x_{q}+\tau)$ for class $c_{neigh}$, 
\[
f_{\theta^{'},c_{neigh}}(x_{q}+\tau)-\underset{k\neq c_{neigh}}{max}f_{\theta^{'},k}(x_{q}+\tau)\geq\gamma_{2}>0
\]
.
\item A null-space or near-null constraint for other images $x_{i}$. For
small perturbations $\parallel\delta\parallel\leq\frac{\gamma_{2}}{2L}$
(from Lemma 2), $(x_{q}+\tau+\delta)$ remains confidently in class
$c_{neigh}$. Any similar $x_{i}(\neq x_{q})$, adding $\tau$ does
not (significantly) alter their logits, unless $x_{i}$ is extremely
close to $x_{q}$, because of null space constraint, i.e. $f_{\theta^{'}}(x)\approx f_{\theta^{'}}(x+\tau)$.
So, they maintain their original class, unless the model explicitly
learns that $x_{i}$ is close enough to $x_{q}$, that flips the class
into $c_{neigh}$. Consequently, the set of top-$k$ mages having
the highest $c_{neigh}$ class scores is unchanged for $\parallel\delta\parallel\leq\frac{\gamma_{2}}{2L}$
(Proof provided in the supplementary).
\end{itemize}

\subsubsection{Lemma 2 ( Margin Bound)}

For an image $x$, being in class $c_{neigh}$ with margin $\gamma_{2}$
means:

\begin{equation}
f_{\theta^{'},c_{neigh}}(x)-\underset{k\neq c_{neigh}}{max}f_{\theta^{'},k}(x)\geq\gamma_{2}>0\label{eq:margin_eq}
\end{equation}

where, $f_{\theta^{'},c_{neigh}}$ is the logit for class $c_{neigh}$
and $f_{\theta^{'},k}$ logit for any class $k\neq c_{neigh}$.

Suppose, we perturb $x$ by a small $\delta$ with $\parallel\delta\parallel\leq\epsilon$.
The network is assumed L-lipschitz in its logit space, i.e. 

\[
\parallel f_{\theta^{'},j}(x+\delta)-f_{\theta^{'},j}(x)\parallel\leq L\parallel\delta\parallel,\forall j
\]
.

We want to see, how this affects the margin in \ref{eq:margin_eq}.

Logit for class $c_{neigh}:$ 
\[
f_{\theta^{'},c_{neigh}}(x+\delta)\geq f_{\theta^{'},c_{neigh}}(x)-L\parallel\delta\parallel
\]

And for other classes : 

\[
f_{\theta^{'},k}(x+\delta)\leq f_{\theta^{'},k}(x)+L\parallel\delta\parallel,\forall k\neq c_{neigh}
\]
.

Defining a new margin at $(x+\delta)$ as 
\[
M_{c_{neigh}}(x+\delta)=f_{\theta^{'},c_{neigh}}(x+\delta)-\underset{k\neq c_{neigh}}{max}[f_{\theta^{'},k}(x+\delta)]
\]

by Lipschitz bound,

\begin{eqnarray*}
M_{c_{neigh}}(x+\delta) & \geq & \left[f_{\theta^{'},c_{neigh}}(x+\delta)-L\parallel\delta\parallel\right]-\left[\underset{k\neq c_{neigh}}{maxf_{\theta^{'},k}(x)}+L\parallel\delta\parallel\right]\\
M_{c_{neigh}}(x+\delta) & = & \left[f_{\theta^{'},c_{neigh}}(x)-\underset{k\neq c_{neigh}}{maxf_{\theta^{'},k}(x)}\right]-2L\parallel\delta\parallel
\end{eqnarray*}

But, from \ref{eq:margin_eq}, we know $f_{\theta^{'},c_{neigh}}(x)-\underset{k\neq c_{neigh}}{max}f_{\theta^{'},k}(x)\geq\gamma_{2}$.

$\therefore$
\begin{eqnarray*}
M_{c_{neigh}}(x+\delta) & \geq & \gamma_{2}-2L\parallel\delta\parallel
\end{eqnarray*}

$\therefore$as long as $2L\parallel\delta\parallel<\gamma_{2}$,
or $\parallel\delta\parallel\leq\frac{\gamma_{2}}{2L}$, $M_{c_{neigh}}(x+\delta)>0$.

That means $(x+\delta)$ remains in the class $c_{neigh}$. 

\subsubsection{Theorem 1 (Trigger-Based Method Has Larger Robustness Radius)}

Let, $\rho_{std}(x_{q})$ be the local retrieval radius (Definition
1) when using the standard feature extractor $f_{\theta}$ and let
$\rho_{trigger}(x_{q})$ be the local retrieval radius when using
the trigger-based method using $f_{\theta^{'}}$. Under the assumptions
of Lemma 1 (margin $\gamma_{2}$, null-space property, Lipschitz bound
$L$) and typical conditions for the standard embedding of $f_{\theta}$,
there exists $\epsilon^{*}>0$ such that : 
\[
\rho_{trigger}(x_{q})\geq\epsilon^{*}>\rho_{std}(x_{q})
\]
In other words, the trigger-based approach guarantees a strictly larger
neighborhood around $x_{q}$ in which the top-$k$ neighbors remain
unchanged.

\paragraph{Proof}
\begin{enumerate}
\item Trigger-Based Radius

By Lemma 1, if the margin $\gamma_{2}$ is enforced for $(x_{q}+\tau)$,
small perturbations $\parallel\delta\parallel\leq\frac{\gamma_{2}}{2L}$
do not alter which images get classified as dummy. Consequently, the
ranking of images by $f_{\theta^{'},c_{neigh}}(x_{q}+\tau)$ does
not change for perturbations up to $\frac{\gamma_{2}}{2L}$. Thus,
$\rho_{trigger}(x_{q})\geq\frac{\gamma_{2}}{2L}$.
\item Standard Radius

In general, standard embedding do not enforce a margin specifically
for $x_{q}$. Suppose there exist at least two database images, $x_{i}$
and $x_{j}$, at nearly the same distance (or similarity) from $f_{\theta}(x_{q})$.
Then even a small $\parallel\delta\parallel$ can reorder $\parallel f_{\theta}(x_{q}+\delta)-f_{\theta}(x_{q})\parallel$
vs $\parallel f_{\theta}(x_{q}+\delta)-f_{\theta}(x_{i})\parallel$.
This can change the top-$k$ neighbor set. Hence, in a typical setting
with many images in the database, we cannot guarantee $\rho_{std}(x_{q})$
is significantly larger than zero. Indeed, without an explicit margin,
one cannot ensure stable top-$k$ retrieval under small perturbations.
\end{enumerate}
Putting these together, we see $\rho_{std}(x_{q})$ can be arbitrarily
small if the embedding is dense or if there is no special margin around
$x_{q}$. Meanwhile, $\rho_{trigger}(x_{q})$ is bounded by $\frac{\gamma_{2}}{2L}$.
Thus, we conclude $\rho_{trigger}(x_{q})>\rho_{std}(x_{q})$, stating
the proposed method is more robust in the noisy environment.

\subsubsection{Self-Retrieval in Presence of OOD Data}

We want to check if the proposed method is still be successful even
if the training set contains OOD samples while retrieving. Eventually
we want to show that with high probability, no OOD data $x_{ood}$
in the training set will have a larger dummy-class probability (after
adding the trigger) than the query $x_{q}$ has. Concretely:

\[
P(y_{c_{neigh}}|f_{\theta^{'}},x_{ood}+\tau)\ngtr P(c_{neigh}|f_{\theta^{'}},x_{q}+\tau)
\]

for all OOD images.
\begin{description}
\item [{Proof:}] 
\[
B_{q}=P(c_{neigh}|f_{\theta^{'}},x_{q}+\tau),
\]
 be the softmax probability score for the dummy class $c_{neigh}$
for a query image $x_{q}$ when added to a trigger. And \\
\[
B_{o}^{i}=P(c_{neigh}|f_{\theta^{'}},x_{ood}^{i}+\tau)
\]
 is the same for OOD samples. We assume each $B_{ood}^{i}$ is a random
variable (due to randomness in the network, noise, or distribution
shifts).
\begin{description}
\item [{Key}] \textbf{Assumption}:
\begin{description}
\item [{i.}] Most OOD inputs produce a diffuse softmax distribution---i.e.,
no single class logit (including the dummy or fake class) dominates.
\item [{ii.}] Concretely, we assume the \textbf{OOD dummy-class }scores
have \textbf{sub-Gaussian} tails around a mean $\mu_{ood}$.\\
Formally, for each $B_{o}^{i}$, $\mathbb{E}[B_{o}^{i}]=\mu_{ood}$,
$B_{o}^{i}$ is $\sigma^{2}$-sub-Gaussian.
\item [{iii.}] On average, the dummy-class probability for OOD samples
is lower than that of query images when added to the trigger, i.e.
$\mu_{ood}<B_{q}$.
\item [{iv.}] Let $M$ be the number of OOD samples in the training set. 
\end{description}
\end{description}
The gap, $\Delta := B_q - \mu_{\text{ood}} > 0$
\begin{description}
\item [{Sub-Gaussian}] \textbf{Bound}
\begin{description}
\item [{i.}] \emph{Bounding a single OOD Sample} 

For a single OOD sample $x_{ood}$, need to estimate the value of
$P(B_{o}\geq B_{q})$. Because $B_{o}$ is $\sigma^{2}$-sub-Gaussian
with mean $\mu_{ood}$, we have (one-sided tail bound):

\[
P\left(B_{o}-\mu_{ood}\geq t\right)\leq\,exp\left(-\frac{t^{2}}{2\sigma^{2}}\right)
\],for all $t>0$.

Here, we choose $t=\triangle=B_{q}-\mu_{ood}>0$. Therefore,
\[
P(B_{o}\geq B_{q})=P\left(B_{o}-\mu_{ood}\geq\triangle\right)\leq\,exp\left(-\frac{\triangle^{2}}{2\sigma^{2}}\right).
\]

\item [{ii.}] \emph{Union Bound Across All OOD Sample}

We want no OOD sample $x_{ood}^{i}$ to exceed $B_{q}$, i.e. $\underset{1\leq i\leq M}{max}B_{o}^{i}<B_{q}.$

By the union bound,

\[
P\!\left(\exists\, i : B_o^{i} \ge B_q \right)
   \le \sum_{i=1}^{M} P\!\left(B_o^{i} \ge B_q \right).
\] 

But each term is bounded by $exp\left(-\frac{\triangle^{2}}{2\sigma^{2}}\right)$.
Therefore, 

\textbf{
\[
P\left(\exists\,i:B_{o}^{i}\geq B_{q}\right)\leq M\times exp\left(-\frac{\triangle^{2}}{2\sigma^{2}}\right)
\]
}
\item [{iii.}] High-probability guarantee
\end{description}
Let denote 

\[
\epsilon=M\times exp\left(-\frac{\triangle^{2}}{2\sigma^{2}}\right)
\]

Then, with probability at least $1-\epsilon$, for all, $i=1,..M.$
\end{description}
Hence, the query $x_{q}$ outperforms all OOD samples on the dummy-class
score with probability $1-\epsilon$. As a result, $x_{q}$ will be
retrieved (or “self-retrieved”) when searching via dummy-class probabilities.
\end{description}

\subsection{Trigger Approximation}

We aim to find a globally orthogonal trigger for the entire data distribution.
As global orthoganality refers to the vector that belong to the null
space of the whole ( inclusion of train and test) data distribution.
However, using only the training data as an approximation of the full
distribution proves insufficient for more challenging datasets, where
test data points can significantly differ from those in the training
set. In such cases, the trigger identified from the training set may
not remain orthogonal for a given query point. Instead, designing
the trigger to be orthogonal at the query point ensures it to remain
locally-consistent. Moreover, our fine-tuning loss function inherently
enforces classifier insensitivity to the trigger. We found that the
negative impact of this adjustment is significantly smaller than the
issue of failing to find an orthogonal trigger at the query point.
Therefore, we adopt the following loss function to optimize for a
query-local orthogonal trigger:

\begin{equation}
\underset{\tau_{q}}{\text{min}}\mathcal{L}\left(f_{\theta}(x_{q}),f_{\theta}(x_{q}+\tau_{q})\right)+\frac{1}{\parallel\tau_{q}\parallel_{F}^{2}}\label{eq:apprx_trig_optimization}
\end{equation}

Once we find the $\tau_{q}$, we also modulate its magnitude with
a factor $\omega$ (in \ref{alg:nearest-neighbour}) to make sure
that the distortion is neither too small and nor too large.

The equation finds a perturbation to be added to the incoming sample
such that the outputs of the original and perturbed samples remain
similar. A trivial solution would be a zero-vector perturbation; however,
the second component of the equation prevents this outcome. Therefore,
the most plausible solution is that the optimizer identifies a vector
which, when projected into the embedded space, exhibits orthogonality
at the location where $x_{q}$ has been projected in the embedding
space.

\section{Experiment}

\subsection{Datasets and Model architecture}

To evaluate performance of our proposed method we use two different
CNN model architectures ResNet-18 and WideResNet50 on four different
foreknown datasets, e.g. MNIST \cite{LeCun1998}, SVHN \cite{Netzer2011},
CIFAR10 \cite{Krizhevsky2009}, and GTSRB \cite{Stallkamp2011}.

\subsection{Baselines}

To evaluate the effectiveness of the proposed approach, we compare
it with conventional distance- and similarity-based nearest neighbour
search methods, including Euclidean ($L_{2}$) distance and cosine
similarity (CS). These baselines were chosen because most nearest
neighbour methods rely on distance or similarity measures. Such metrics
can be applied in either raw pixel space or deep feature space. However,
prior studies and empirical evidence show that neighbour retrieval
in pixel space is highly sensitive to perturbations, often failing
under even minor noise.

To address this limitation, we borrow DNN's to extract feature representations,
then The $L_{2}$ and CS metrics are employed in the resulting feature
space to determine neighborhood relationships. We have selected penultimate
layer of DNN to extract the feature. The choice of this layer was
deliberate for several reasons: (a)hidden layers have very high dimensionality,
which increases computational overhead; (b) hidden layers may not
capture semantic similarity effectively; and (c) their features can
be highly entangled, making similarity measures less meaningful.

\subsection{Trigger-based NN setting}

\subsubsection{Fine-tuning setting}

To train Targeted Manifold Manipulation-Nearest Neighbour (or TMM),
we fine-tune fully connected (FC) layer of the pre-trained model,
$f_{\theta}$, for 1 epoch. To avoid accidental catastrophic forgetting,
Elastic weight consolidation (or EWC) \cite{Kirkpatrick2017} loss
is included with the original TMM loss function. Learning rate $\alpha_{ftrain}=0.001$.
Adam has been used as optimizer, with the batch size of $256$.

\subsubsection{Trigger generation}

Trigger optimization was capped at $300$ iterations, though convergence
was typically reached within $100$ iterations with a fixed learning
rate of $0.015$. And empirically found that setting the trigger intensity
$\omega$ to the standard deviation of $x_{q}$ provides optimal performance.
We generate a separate trigger for each query image.

\subsection{Self-Retrieval}

We attempt to retrieve a few samples (i.e. query samples) from a search
space when the same instance being a part of the search space ( i.e.
$\mathcal{D}_{search}$). The motivation is to examine if the proposed
method truly is capable of capturing neighbourhood of any given query
sample. If so, TMM-NN should return$1^{st}$ NN as the query sample
itself. The retrieval percentage measures the success rate when the
method correctly identifies the query sample itself as its $1^{st}$
nearest neighbor.

\begin{figure*}
\subfloat[\label{fig:bright_xq_train}Self-retrieval performance under varying
brightness levels of query samples . The comparison includes two baselines
($L_{2}$-distance and cosine similarity (CS)) and our proposed method.]{\begin{centering}
\includegraphics[width=3.3cm,totalheight=3.3cm]{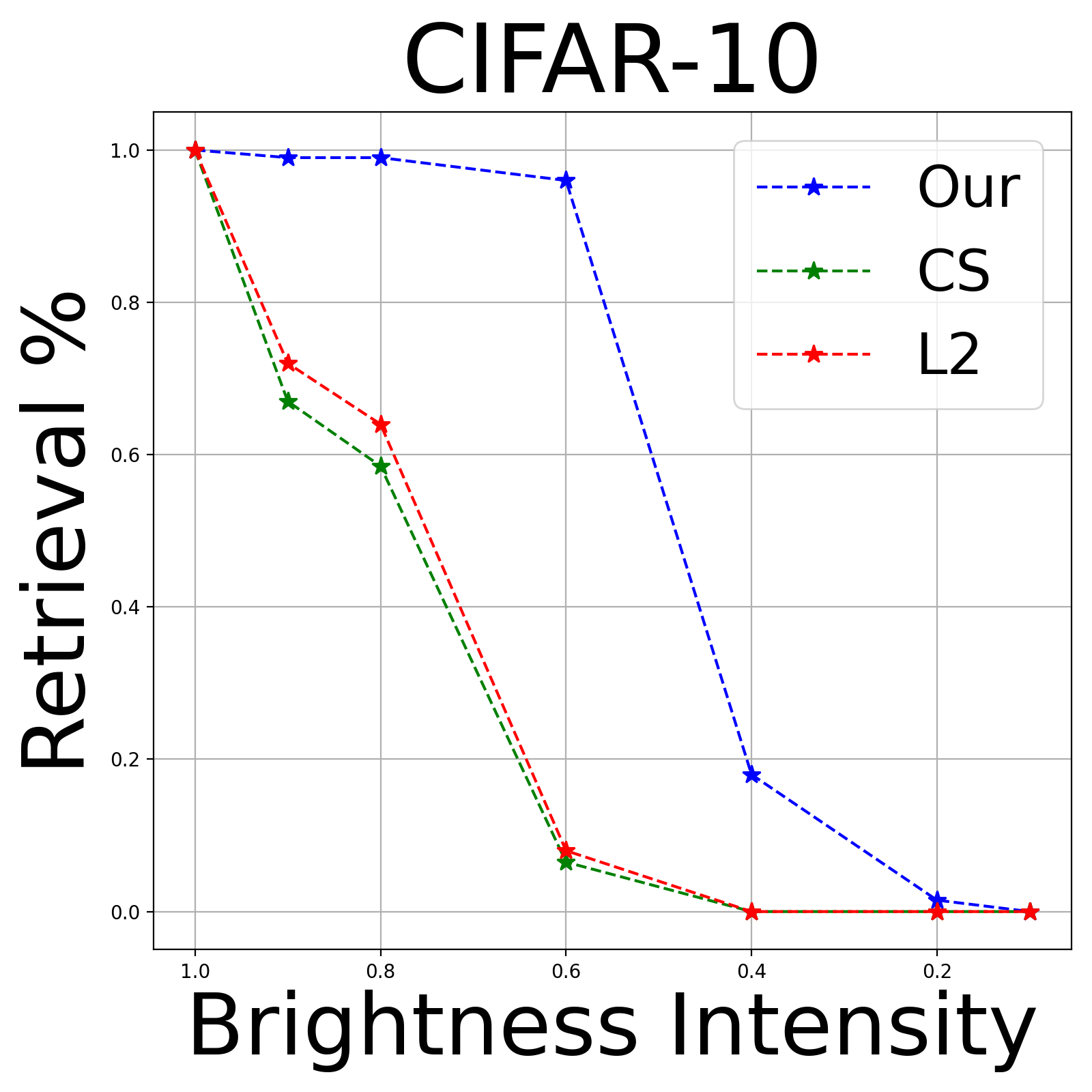}\includegraphics[width=3.3cm,totalheight=3.3cm]{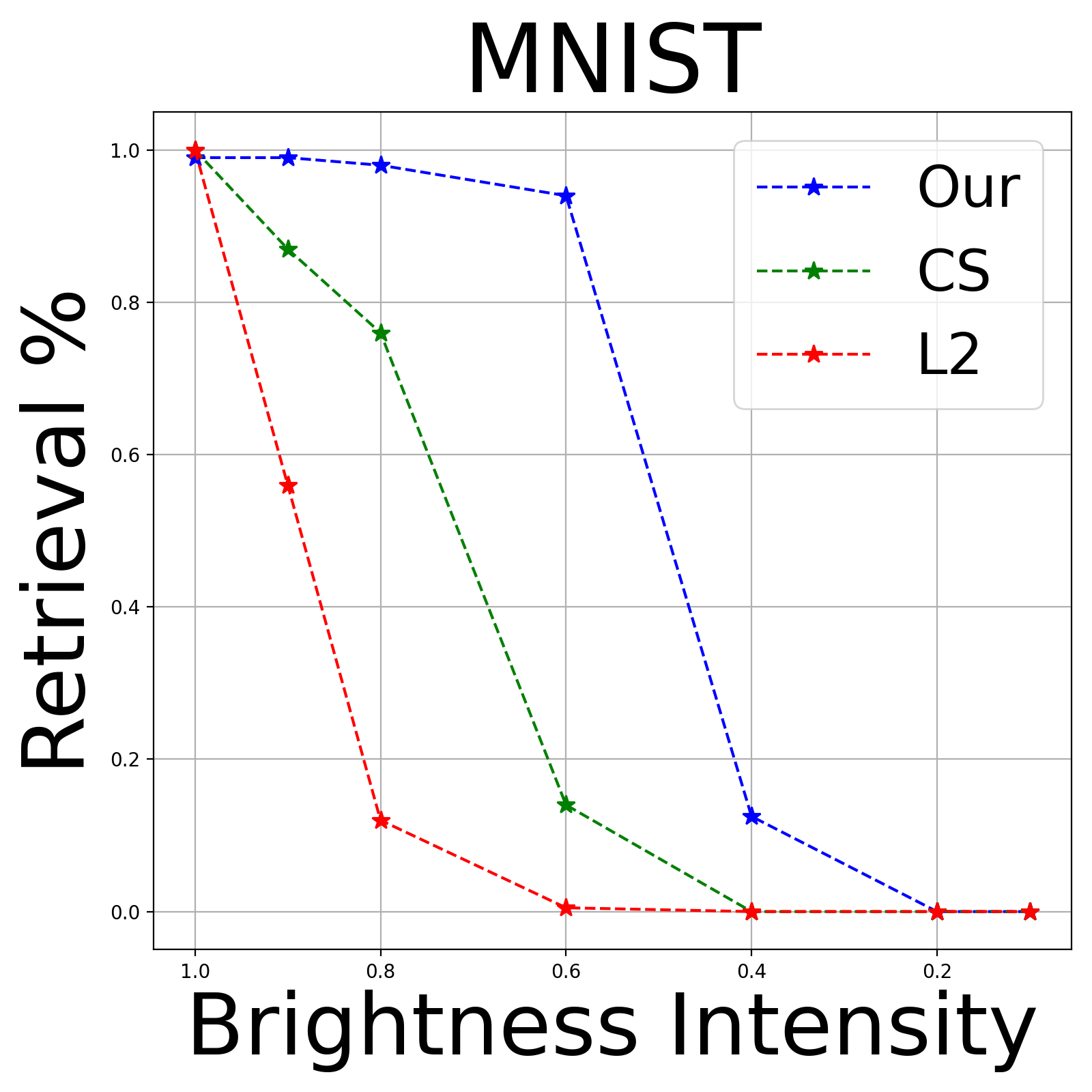}\includegraphics[width=3.3cm,totalheight=3.3cm]{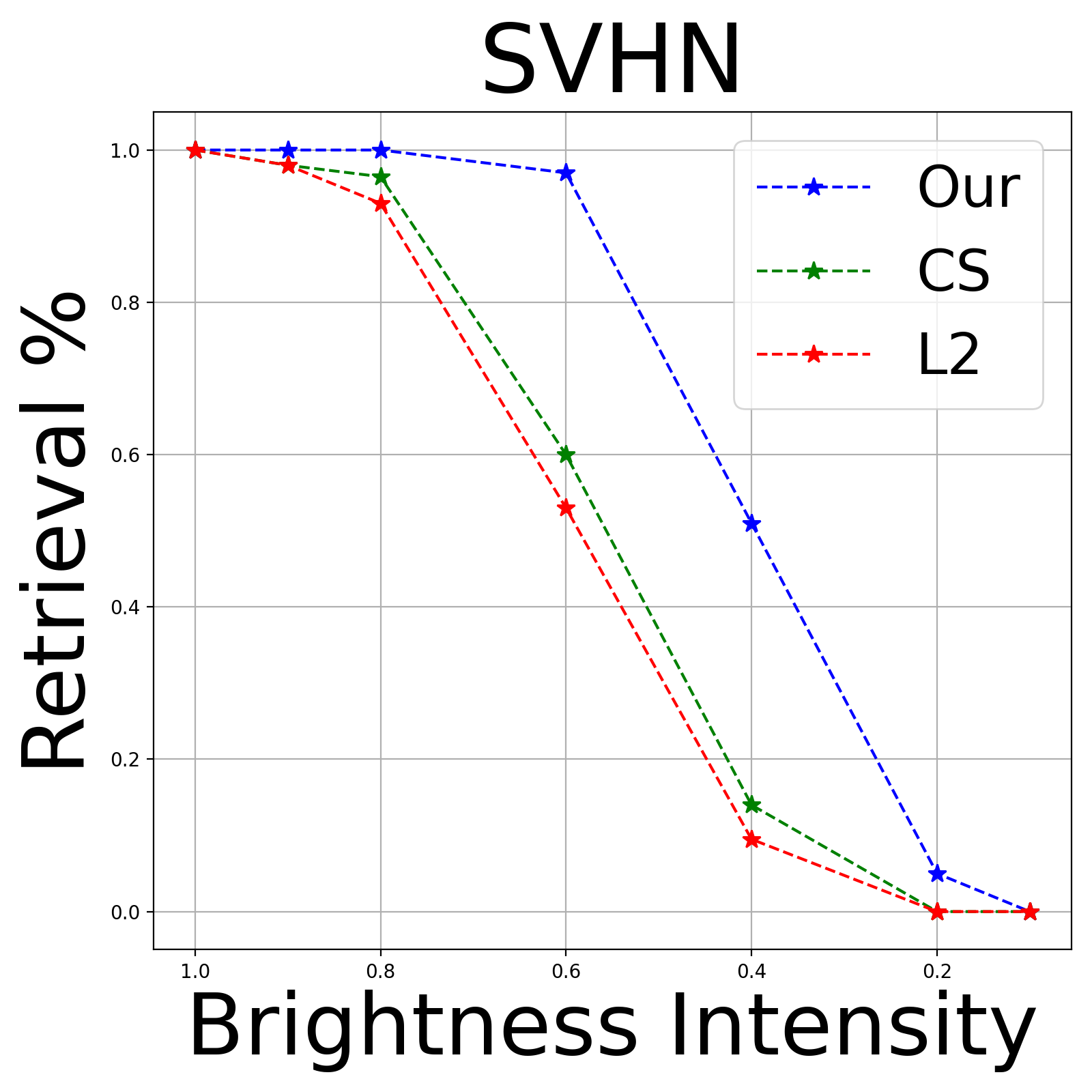}\includegraphics[width=3.3cm,totalheight=3.3cm]{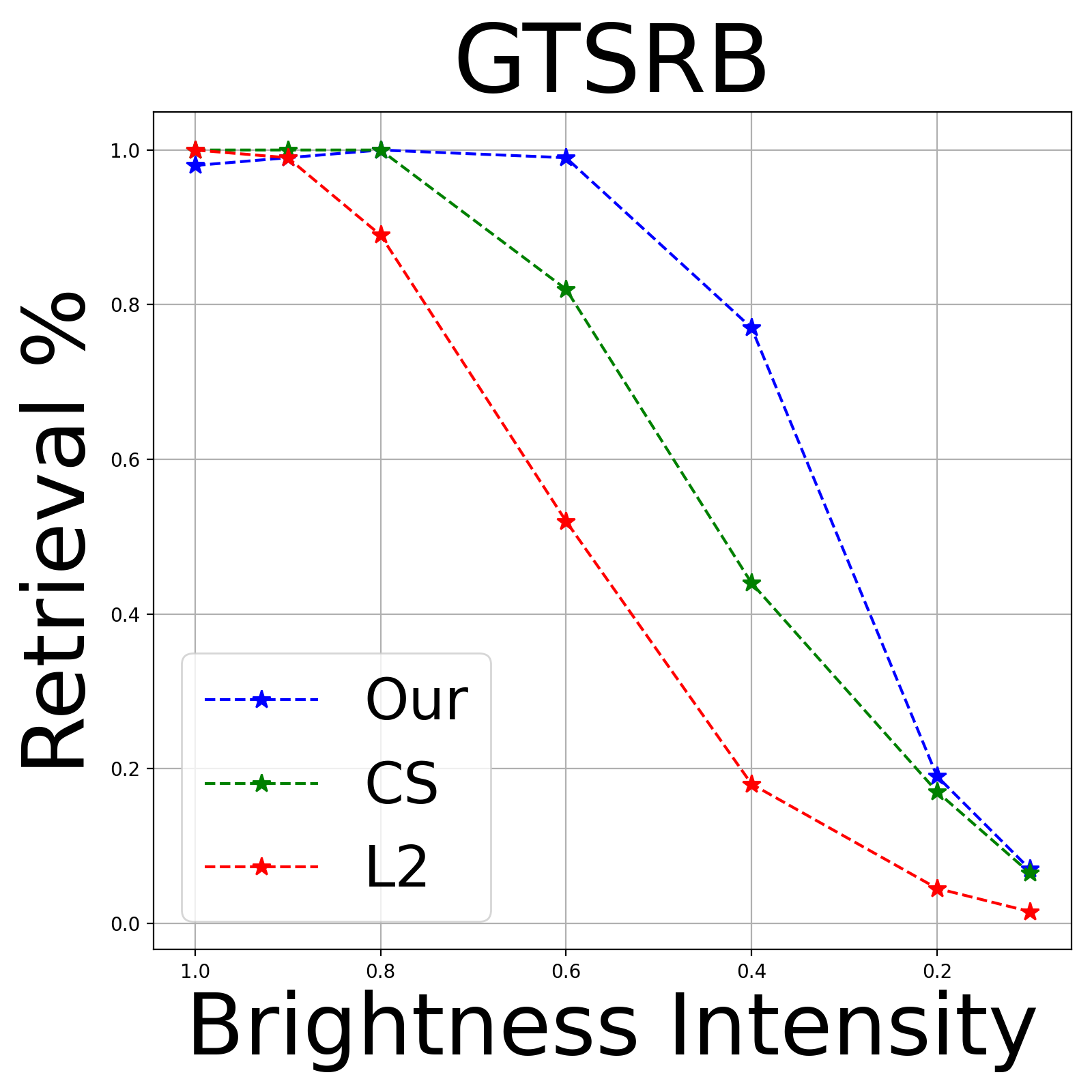}
\par\end{centering}
}

\subfloat[\label{fig:gauss_xq_train}Self-retrieval performance under varying
levels of Gaussian noise (controlled by noise norm). The comparison
includes two baselines ($L_{2}$-distance and cosine similarity (CS))
and our proposed method..]{\begin{centering}
\includegraphics[width=3.3cm,totalheight=3.3cm]{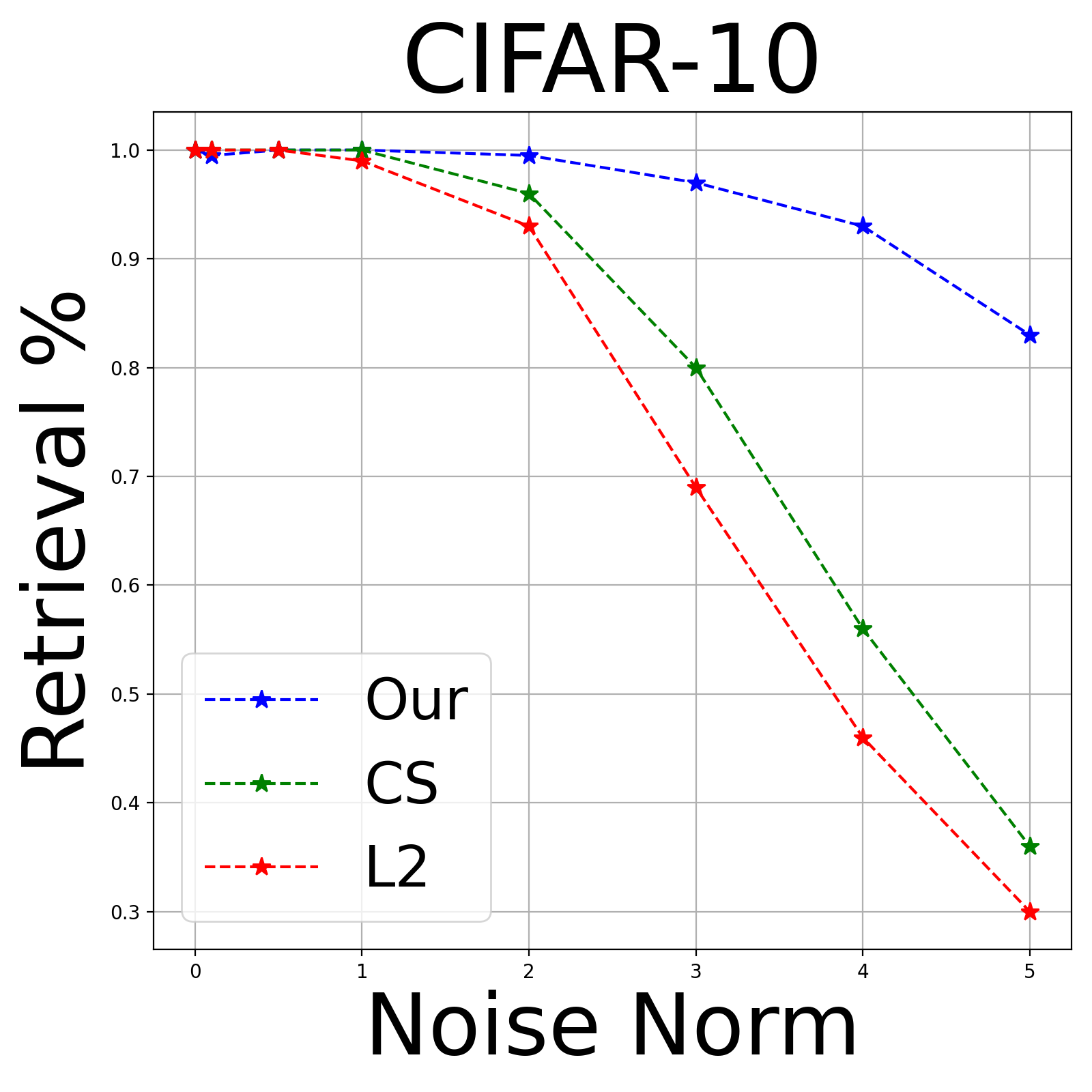}\includegraphics[width=3.3cm,totalheight=3.3cm]{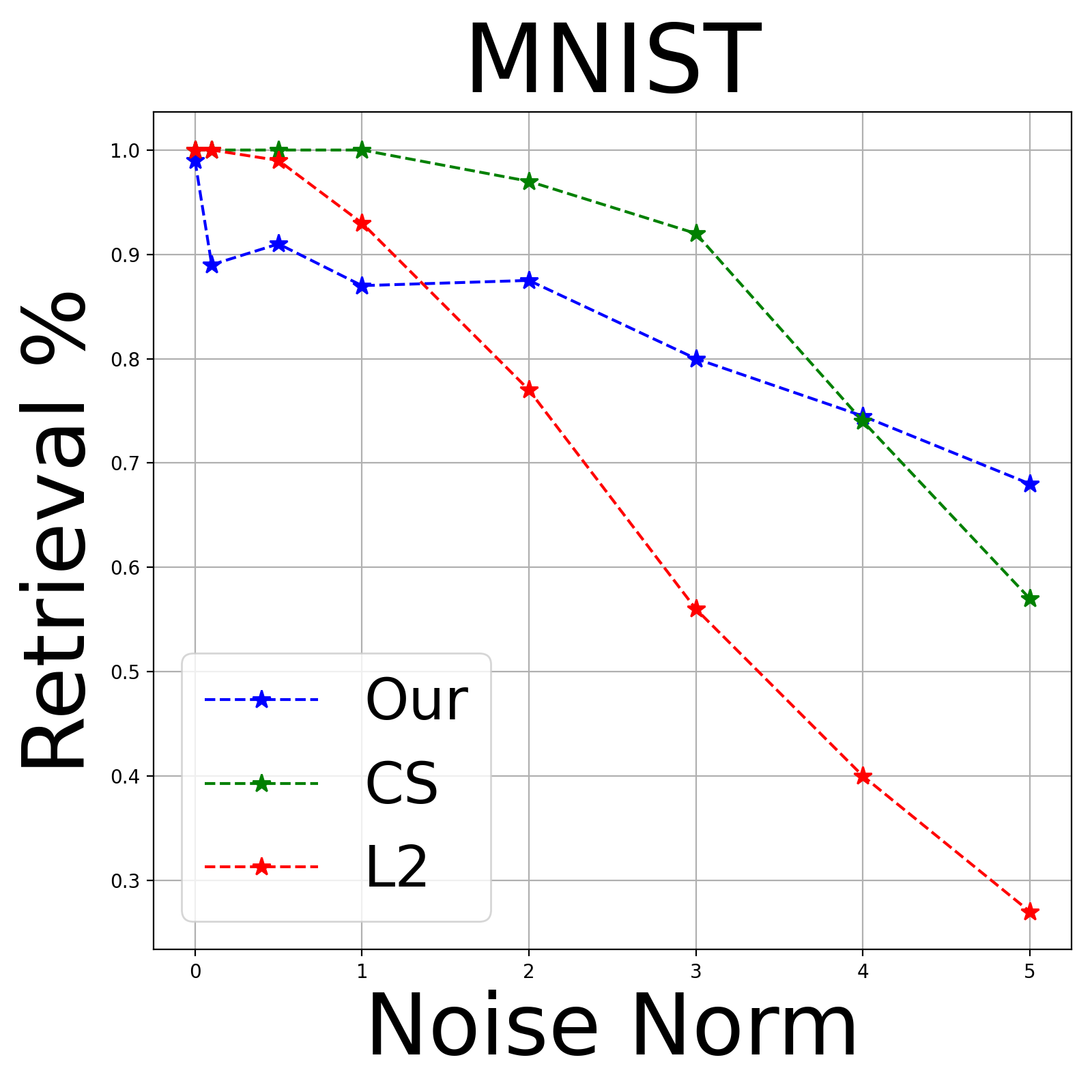}\includegraphics[width=3.3cm,totalheight=3.3cm]{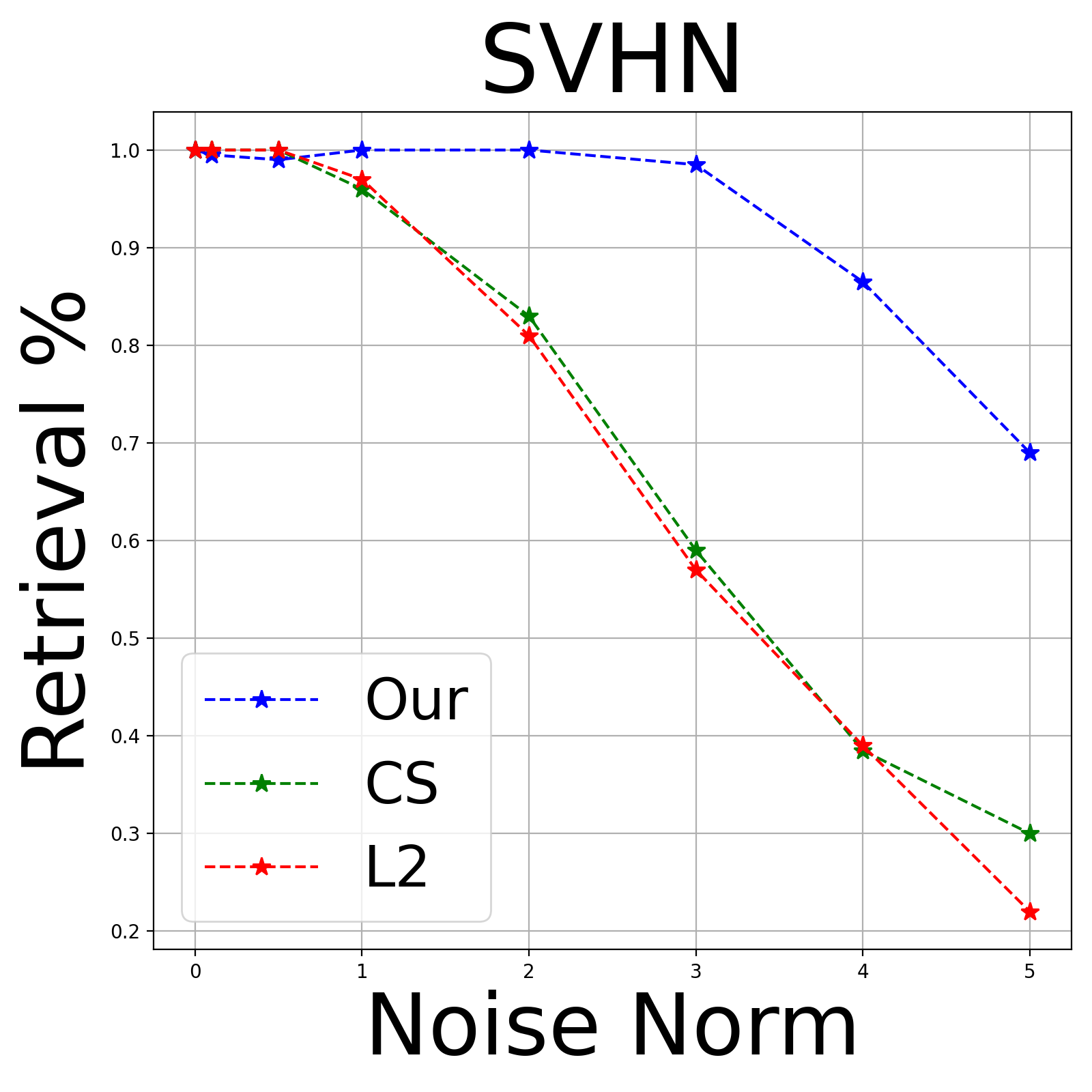}\includegraphics[width=3.3cm,totalheight=3.3cm]{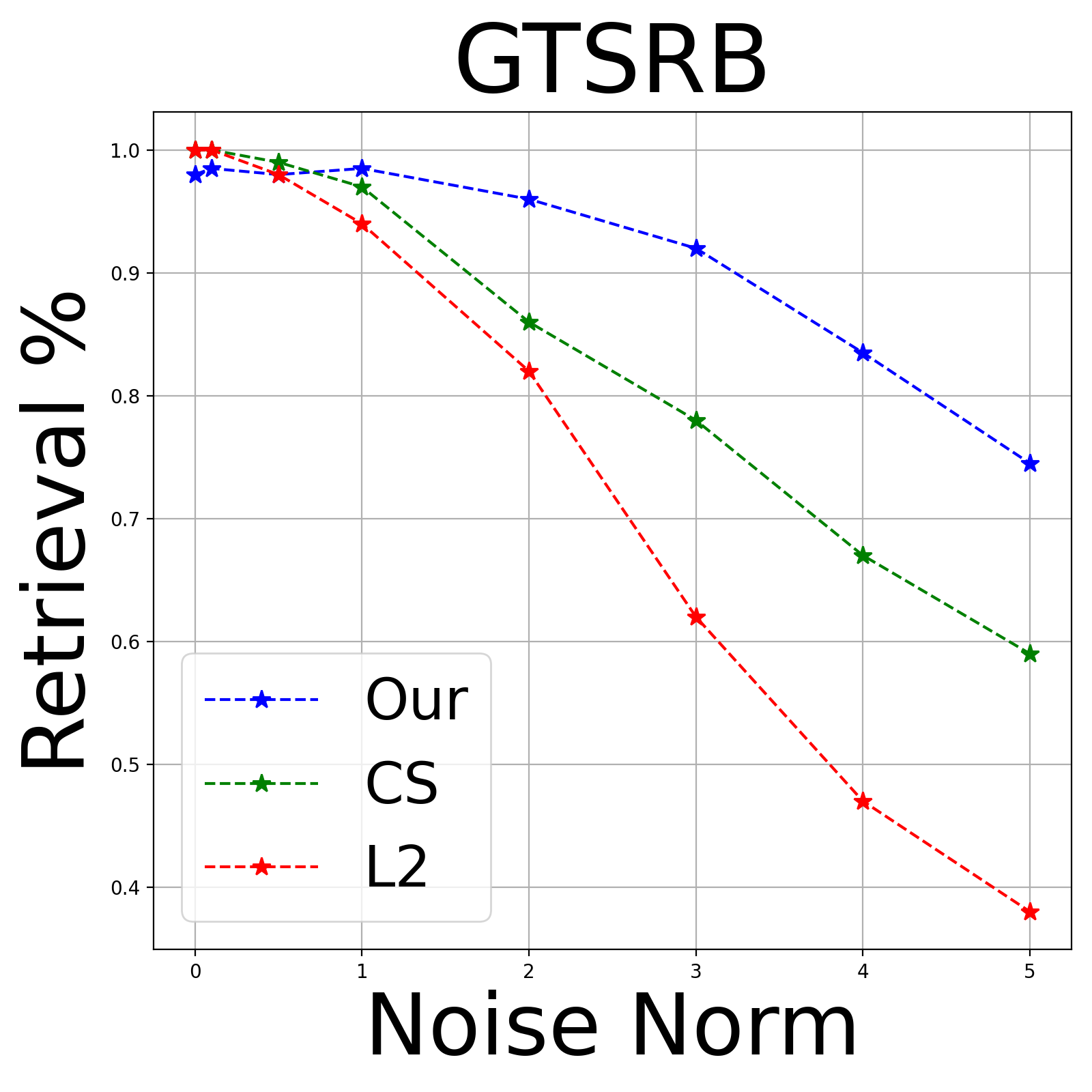}
\par\end{centering}
}\caption{Performance comparison of retrieval robustness under two perturbations---brightness
variation and Gaussian noise---across four datasets (CIFAR-10, MNIST,
SVHN, and GTSRB). In both scenarios, our proposed method consistently
outperforms the baselines ($L_{2}$-distance and cosine similarity
(CS)), demonstrating superior robustness to input distortions.}\label{fig:controlled_experiment}

\end{figure*}

Moreover, we are focused on evaluating the robustness of the retrieval
process. To assess this, we introduce various types of noise at different
intensity levels to the query samples and examine whether the nearest
neighbors remain consistent.
\begin{enumerate}
\item \textit{Self-retrieval }: We form a set of query instances of 200
samples, $X_{q}=\left\{ x_{q}^{i}\right\} _{i=1}^{200}$, randomly
selected form training dataset, where $x_{q}^{i}\in\mathcal{D}_{train}$,
and $\mathcal{D}_{search}=\mathcal{D}_{train}$. Expecting the retrieved
first nearest neighbours are $x_{q}^{i}$ themselves by all the retrieval
methods. In the Figures \ref{fig:bright_xq_train}\ref{fig:gauss_xq_train},\ref{fig:gauss_xq_train},,
the first data points in each graph represent the self-retrieval performance
of the baselines and the proposed method. The results indicate that
while the proposed method performs comparably to the baselines under
normal conditions, the introduction of noise reveals a different outcome.
\item \emph{Robustness for self-retrieval} : We introduce controlled disturbance
(or noise) in the queries, but leaving search dataset as it is. Addition
of noise to the queries as follows -
\begin{enumerate}
\item Brightness Adjustment : We modify the brightness of $x_{q}^{i}$ using
a brightness intensity factor $t_{b}$, where $0.1<t_{b}\leq1$, $1$
being the $100\%$ brightness.
\item Gaussian Noise Addition : We add Gaussian noise to the query samples,
described as i.e. $x_{q}+\triangle x$, where $\parallel\triangle x\parallel_{2}\leq\varepsilon_{g}$
, $0<\varepsilon_{g}\leq5$.
\end{enumerate}
\end{enumerate}

Fig. \ref{fig:controlled_experiment} shows the performance comparison
of the proposed and baselines. From the figs \ref{fig:bright_xq_train},
\ref{fig:gauss_xq_train}, we can see that proposed method along with
the baselines perform similarly when the condition is ideal, i.e.
queries are in benign condition, however, under noise TMM-NN demonstrates
more stable retrieval compared to the baseline methods.

\subsection{Semantic feature preservation}

\begin{figure}
\centering
\includegraphics[width=8.5cm,totalheight=5cm]{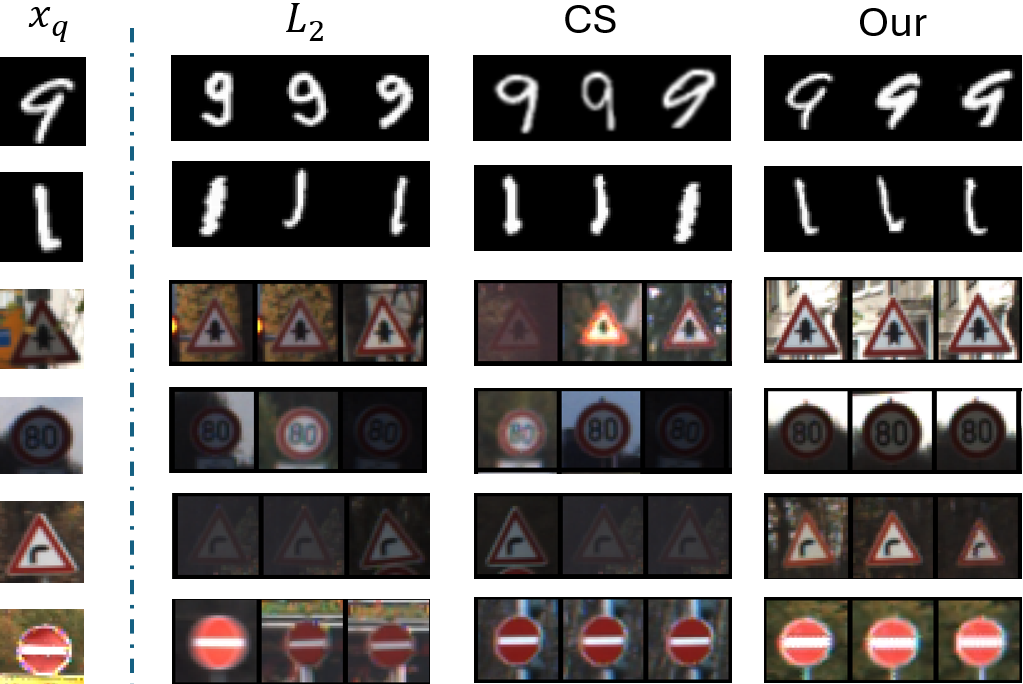}\caption{Illustration of query samples ( $x_{q}$) along with their top-3
nearest neighbours retrieved using two baseline methods ($L_{2}$-distance
and cosine similarity (CS)) and our proposed approach.}\label{fig:Few-nn-examples}
\end{figure}

Fig. \ref{fig:Few-nn-examples} illustrates a few examples of $x_{q}\in\mathcal{D}_{test}$,
along with their top three nearest neighbors retrieved on the GTSRB
and MNIST datasets. The neighbors retrieved by the proposed method
demonstrate a stronger semantic alignment with the query samples.

In the first row, the query sample is the digit “9” from the MNIST
dataset. Notably, the writing style of “9” in the query closely resembles
that of the top three nearest neighbors identified by TMM-NN. Similarly,
other examples show a consistent pattern. TMM-NN not only focuses
on the primary object but also considers the background. In the other
examples, TMM-NN successfully retrieves neighbors with both similar
objects and closely matching backgrounds.

\subsection{LVLMs as Oracle}

In the absence of an established ground truth, human judgment is frequently
regarded as the gold standard. However, obtaining expert annotations
is both costly and time-intensive. As an alternative, we propose leveraging
Large Vision Language Models (LVLMs), whose reasoning capabilities
are often comparable to those of humans, to assess semantic similarity. 

As for each query instance, we generate three sets of k-nearest neighbors
using distinct methods, $L_{2}$, CS, and TMM-NN. We then task an
LLM with determining which neighbor set exhibits the highest degree
of semantic alignment with the query.

For example, we consider the first row from fig. \ref{fig:Few-nn-examples},
where we have query image, handwritten ``9'', and three sets of
different NNs. It is clearly visible that the NNs retrieved by the
proposed method is semantically aligned with the query, and $L_{2}$
and Cosine similarity fail to retrieve such samples as neighbours.
So, we ask GPT-4o with this following prompt - 

\textbf{\emph{Prompt}}\textbf{ : }`` \emph{You are provided with
four files. First one is the query image. And last three are the sets
of nearest neighbour found by $L_{2}$, Cosine, and proposed method
respectively. Can you find which set is the best visually suitable
to the query image? Please make the reasoning in short. And describe
why it is better than the other.}''

\begin{figure}
\centering
\begin{centering}
\includegraphics[width=7cm,totalheight=7cm]{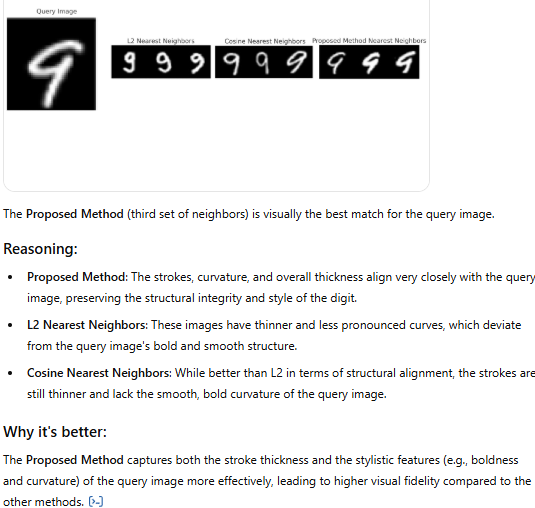}
\par\end{centering}
\caption{GPT-4o response regarding similarity between the query image and
retrieved nearest neighbours.}\label{fig:GPT-model-4o}

\end{figure}

Fig \ref{fig:GPT-model-4o} shows the response received from the GPT-4o,
which seemed reasonable-

\begin{table}
\begin{centering}
\begin{tabular}{ccccc}
\toprule 
\multirow{2}{*}{{\footnotesize Dataset}} & \multicolumn{2}{c}{{\footnotesize ResNet-18}} & \multicolumn{2}{c}{{\footnotesize WideRes50}}\tabularnewline
\cmidrule{2-5}
 & {\footnotesize GPT-4o} & {\footnotesize Gemini} & {\footnotesize GPT-4o} & {\footnotesize Gemini}\tabularnewline
\midrule 
{\footnotesize CIFAR-10} & {\footnotesize 78.50} & {\footnotesize 91.50} & {\footnotesize 89} & {\footnotesize 93.50}\tabularnewline
\midrule 
{\footnotesize SVHN} & {\footnotesize 76.50} & {\footnotesize 87} & {\footnotesize 88} & {\footnotesize 94.00}\tabularnewline
\midrule 
{\footnotesize MNIST} & {\footnotesize 95.5} & {\footnotesize 89.50} & {\footnotesize 83.5} & {\footnotesize 90.50}\tabularnewline
\midrule 
{\footnotesize GTSRB} & {\footnotesize 94.00} & {\footnotesize 95} & {\footnotesize 91.5} & {\footnotesize 97.00}\tabularnewline
\bottomrule
\end{tabular}\caption{Percentage of times LVLM found Nearest Neighbours from TMM-NN to
be better than the baselines.}\label{tab:retrival_test_samples}
\par\end{centering}
\end{table}

Along with GPT-4o, Gemini-1.5 is also being used in this experiment.
By calling api of these models, we compared how much better our method
is than the baselines. We randomly select $200$ samples from the
test dataset for comparison. The percentage of successful cases where
LVLM identifies our neighbors as the most visually appropriate is
recorded in Table \ref{tab:retrival_test_samples}. This clearly show
the superiority of our method in retrieving most semantically similar
nearest neighbors. 

\subsection{Retrieval with Vision Transformer}

\begin{figure}
\begin{centering}
\includegraphics[width=5cm,totalheight=5cm]{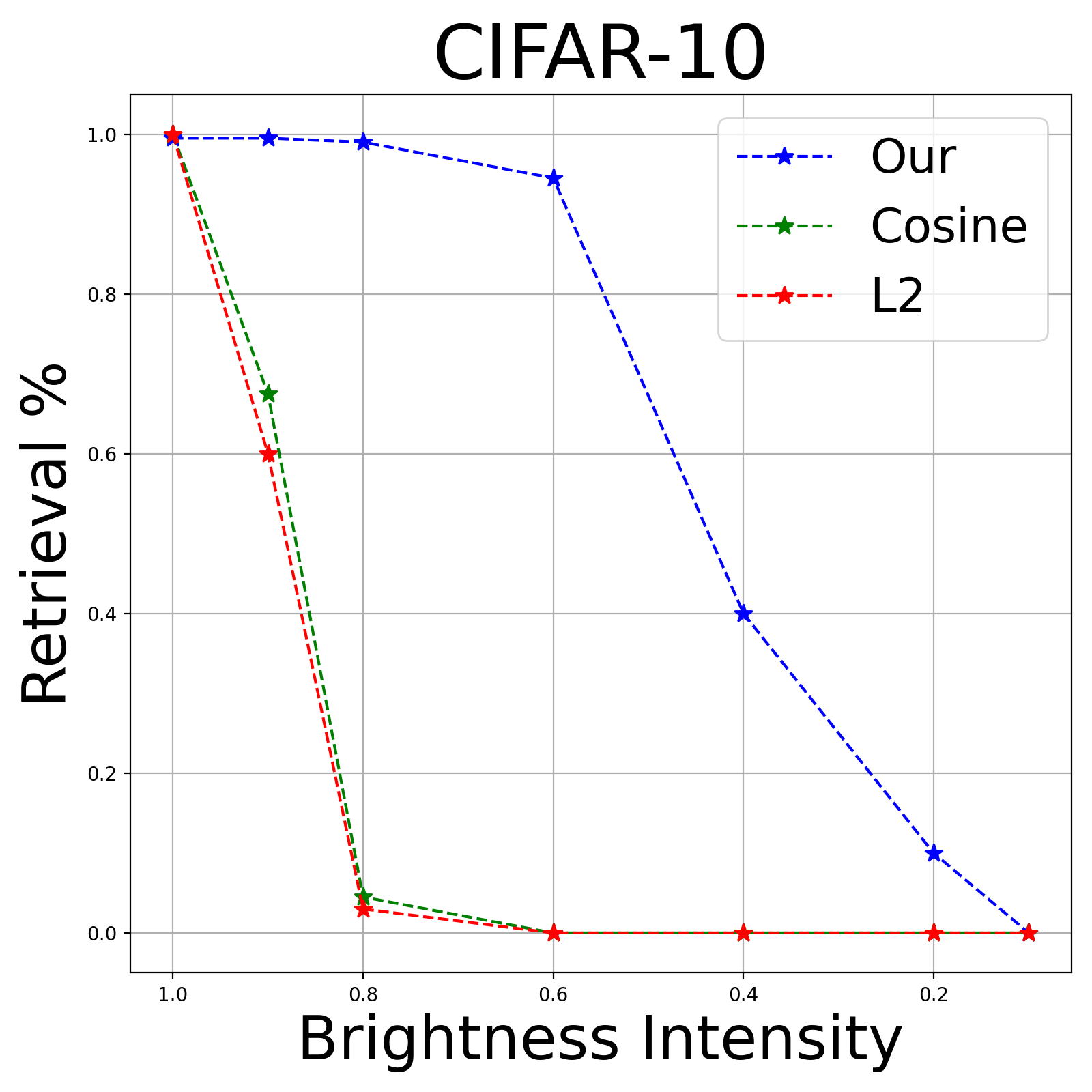}
\par\end{centering}
\caption{Robustness (brightness change) comparison against brightness change
on CIFAR-10 for a ViT model.}\label{fig:Robustness-vit}

\end{figure}

We train a simple ViT on CIFAR-10 dataset and retrieve nearest neighbours
for $200$ randomly selected images as queries under self-retrieval
setting but under brightness change as the query perturbation. From
Figure \ref{fig:Robustness-vit}, the result indicates that the proposed
detection method is not limited by the model architectures, as well
as, it handles self-retrieval when brightness change imposed on the
query samples, indicates robustness.

\subsection{Ablation Study}

In the ablation study we discuss about other two trigger settings
along with different fine-tuning methods to form TMM-NN. Below discuss
studies are done on the CIFAR-10 dataset with ResNet-18 model, under
self-retrieval setting with brightness change as the query perturbation. 

\begin{figure*}
\begin{centering}
\subfloat[\label{fig:Comparison-of-triggers}Different trigger settings .]{\centering{}\includegraphics[width=4cm,totalheight=4cm]{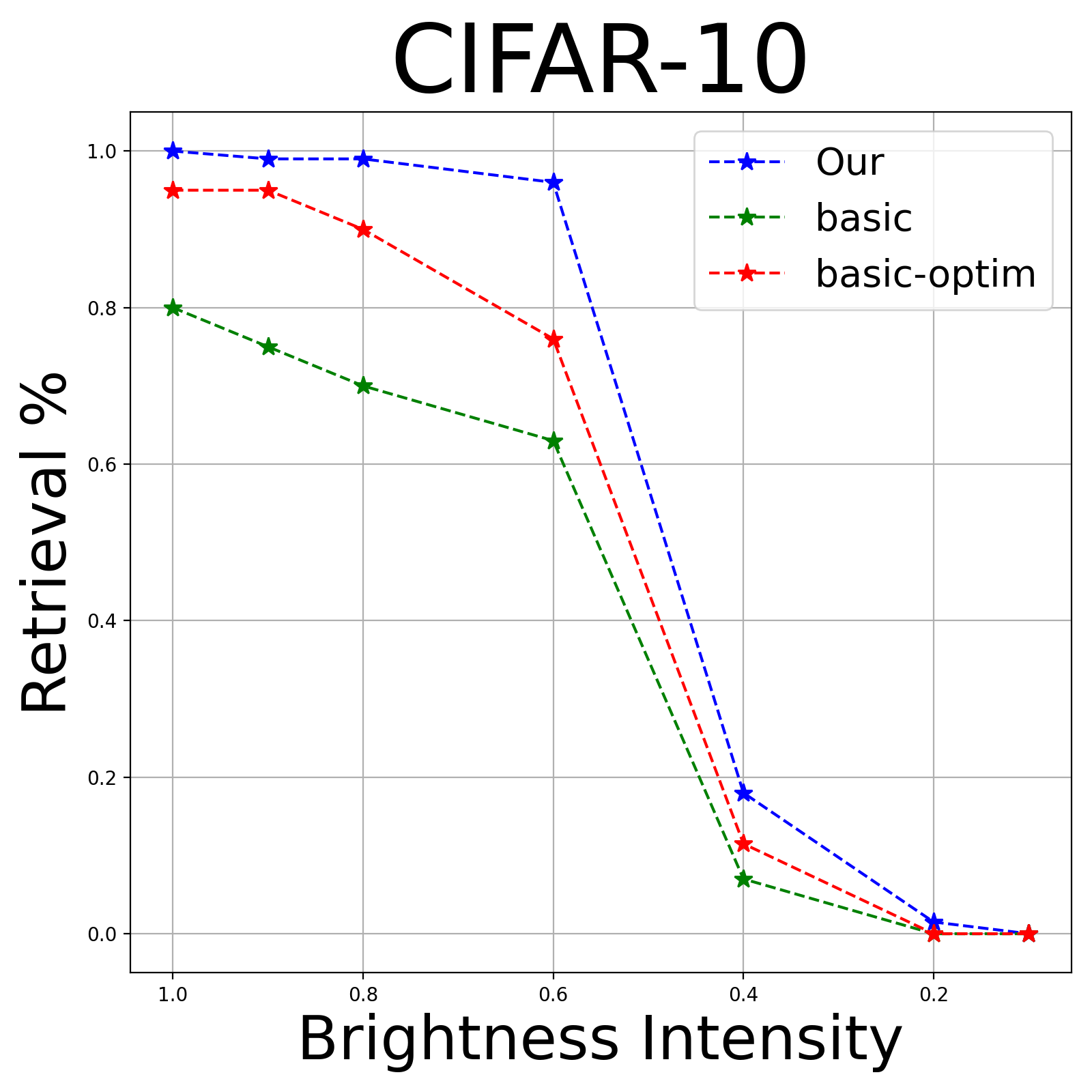}}~~\subfloat[\label{fig:Performance-comparison-layers}Choice of different layers
are for fine-tuning.]{\begin{centering}
\includegraphics[width=4cm,totalheight=4cm]{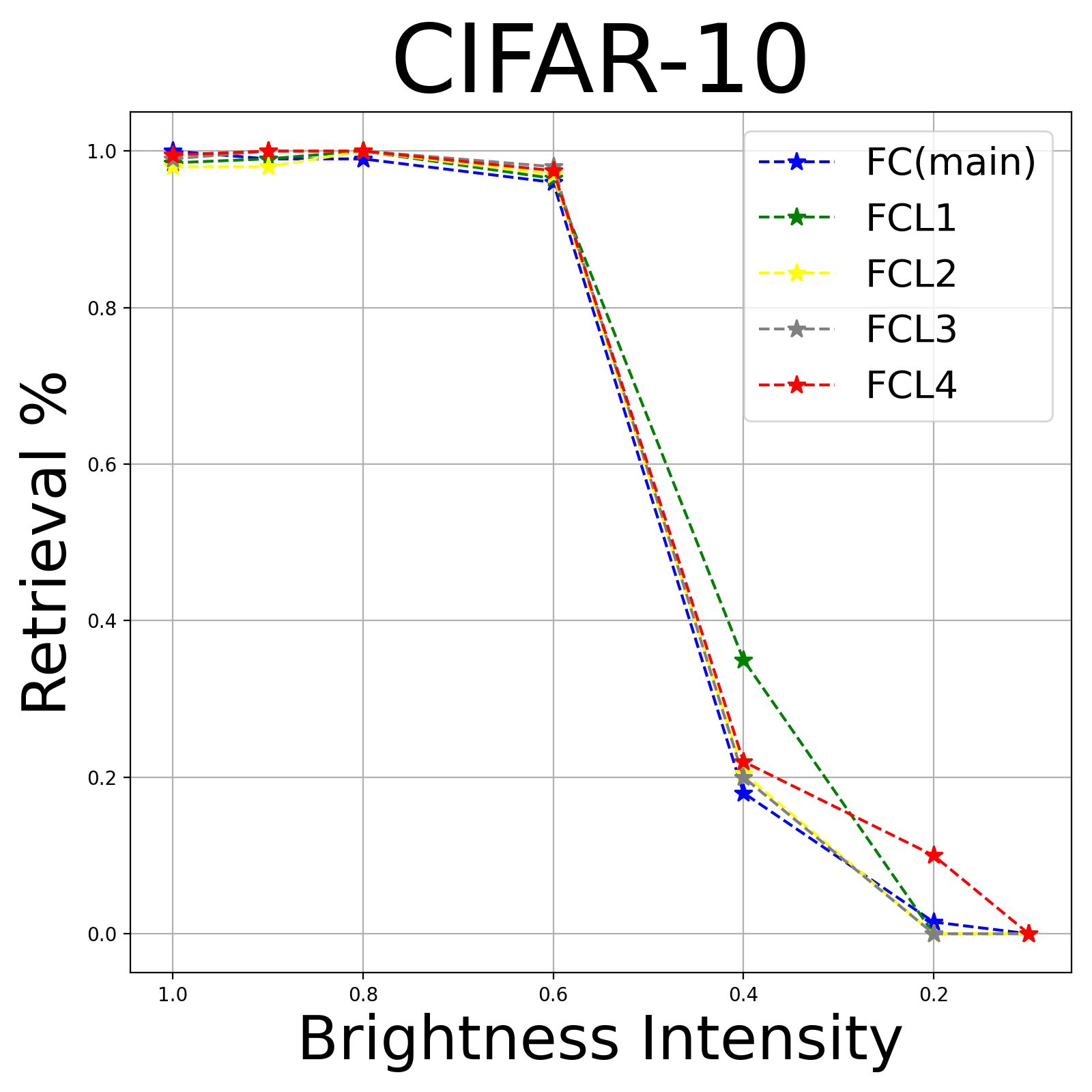}
\par\end{centering}
}1st~~\subfloat[\label{fig:Performance-epochs} Different choice of \#epochs for fine-tuning
.]{\begin{centering}
\includegraphics[width=4cm,totalheight=4cm]{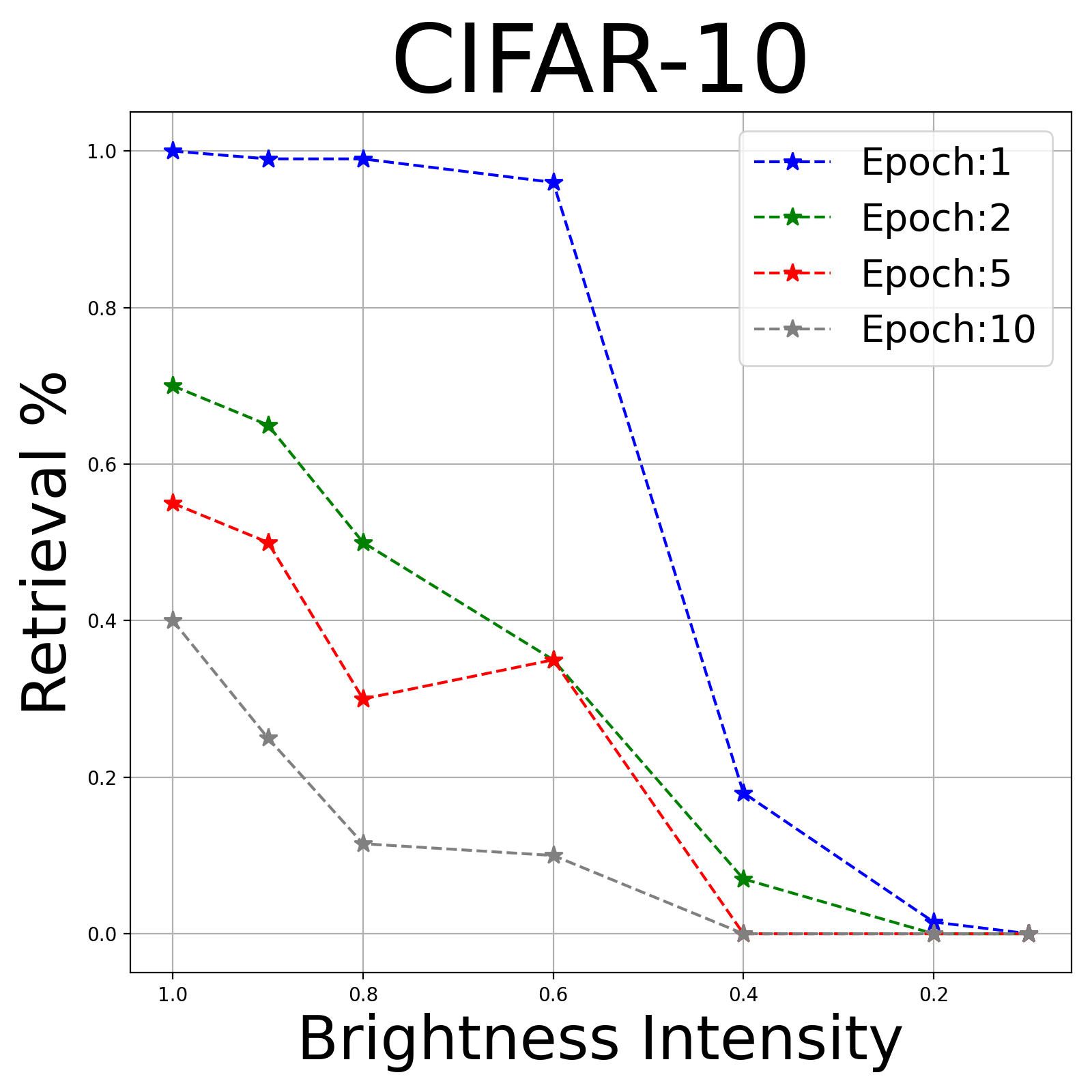}
\par\end{centering}
}
\par\end{centering}
\caption{Performance of self-retrieval against different ablation setting on
CIFAR-10 dataset.}
\end{figure*}

\subsubsection{Different Triggers}

We try other two different trigger settings- 
\begin{figure}[H]
\centering{}\includegraphics[width=6cm,totalheight=6cm]{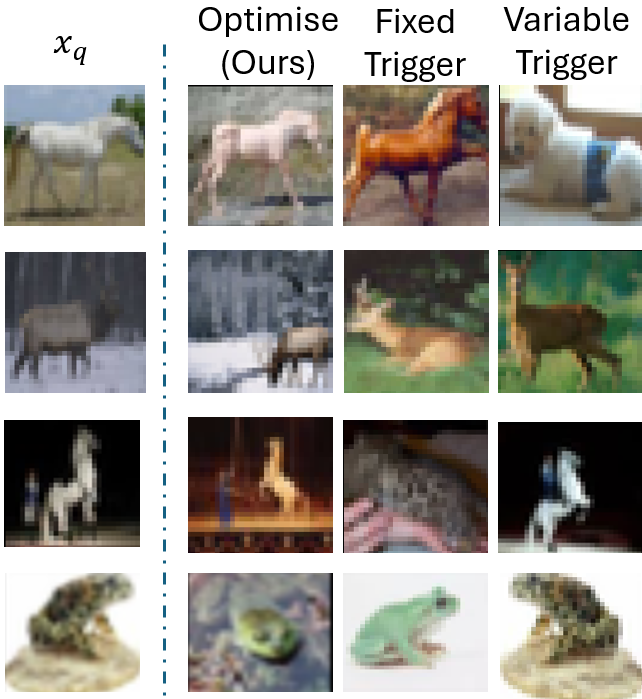}\caption{The query and its $1^{st}$ neighbour when different trigger setting
( fixed trigger, variable trigger, and original proposed trigger)
are used for retrieval process on test data, example provided from
CIFAR-10 dataset.}\label{fig:diff-trigger-nn}
\end{figure}

\begin{enumerate}
\item \emph{Fixed trigger} ($T_{1})$ : a fixed trigger is used at the left
top corner of images to perform the neighbour search. Trigger shape
we used is $3\times3$.
\item \emph{Variable trigger} ($T_{2})$ : a variable trigger values are
used at the left top corner of images. The pixel values are optimized
as we do in the method section.
\end{enumerate}
Figure \ref{fig:Comparison-of-triggers} shows the performance comparison
between the original proposed trigger method (named as ``optimise'')
and above mentioned other two trigger settings. It can be noticed
that the full optimized trigger is superior. Figure \ref{fig:diff-trigger-nn}
shows how retrieved nearest neighbours varies depending on the trigger
setting. But, as long as trigger is maintained orthogonal or out-of-distribution
of the training set, the retrieval method works fairly, as shown in
the above figure. 

\subsubsection{Fine-tune settings}

Fine-tune method adapts a few different setting like 
\begin{enumerate}
\item \emph{Different layer} : Our main experiment considers fine-tuning
the FC layer only. We would like to explore how different layer involvement
during the fine-tuning impact the retrieval performance. We choose
all four layers (Layer 1,2,3,4) one at a time combing with FC layer
during fine-tuning process. We always include the FC layer in the
each combination set as we need to learn the new class i.e. $c_{neigh}$.
From the Fig. \ref{fig:Performance-comparison-layers}, we can state
that the involvement of the different layers does not impose huge
impact on the performance, however, it have been seen than when Layer
4 is trained with FC layer provide slightly better result. Figure
\ref{fig:NN-diff-layes} illustrates retrieved neighbours when different
layers have been selected while fine-tuninig. The first column corresponds
to the original proposed method, which uses only a fully connected
(FC) layer without incorporating any convolutional layers. We can
empirically conclude that the use of FC layer works best on average.
\begin{figure}[H]
\centering
\centering{}\includegraphics[width=6cm,totalheight=6cm]{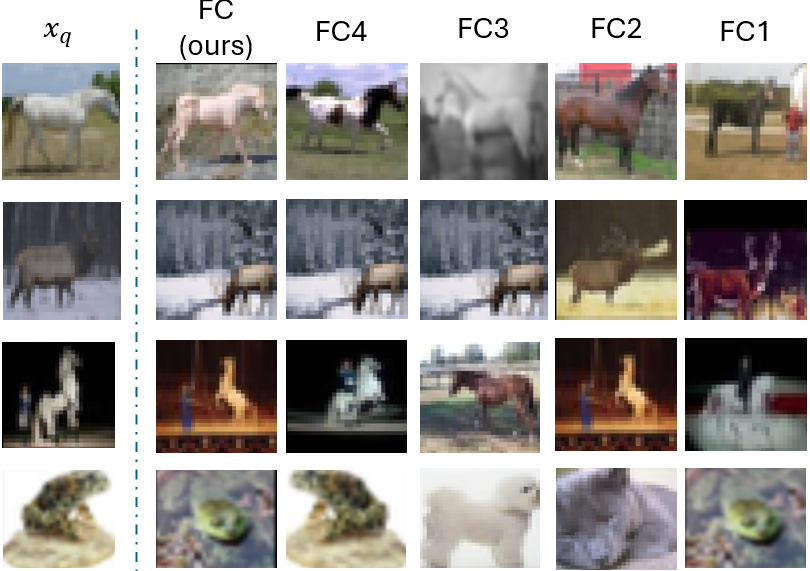}\caption{The query and its $1^{st}$ neighbour when different layers fine-tuned.}\label{fig:NN-diff-layes}
\end{figure}
\item \emph{\# epoch} : Fig. \ref{fig:Performance-epochs} demonstrate that
the higher fine-tune epoch degrades search performance. Higher number
of epoch may change the classifier manifold more thoroughly, resulting
in the poor performance. Figure \ref{fig:nn-epochs} shows the retrieved
NNs for different epochs for fine-tuning.
\begin{figure}
\centering
\centering{}\includegraphics[width=6cm,totalheight=6cm]{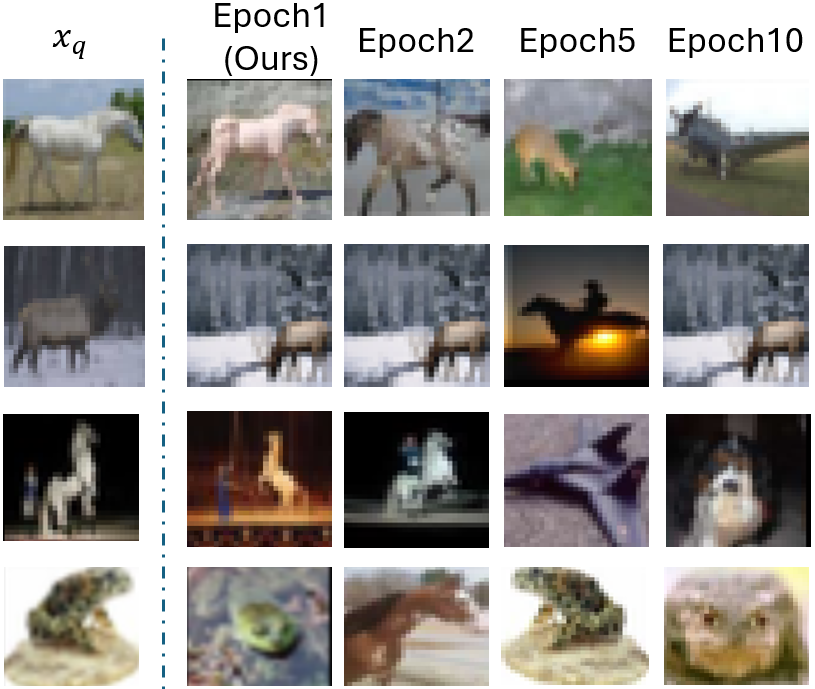}\caption{Evolution of nearest neighbour retrievals across fine-tuning epochs.
The first column shows query samples ($x_{q}$), followed by their
nearest neighbours retrieved at different stages (Epoch 1 with our
method, and Epochs 2, 5, and 10 after further fine-tuning). Our method
(Epoch 1) retrieves semantically consistent neighbours, while increasing
fine-tuning causes the baseline to drift toward less relevant samples.}\label{fig:nn-epochs}
\end{figure}
\end{enumerate}

\subsection{Limitation}

TMM-NN captures semantically aligned samples when asked to search
for neighbours and outperforms other baselines. However, it requires
final layer fine-tuning and thus will generally be more computationally
expensive. Additionally, trigger design is a key step and any misstep
on that can greatly affect the quality of the results. We tested the
robustness of the retrieval against naturally occurring noise, including
brightness changes and Gaussian noise. Additionally, we evaluated
robustness against adversarial perturbations; however, similar to
the baselines, our method failed to retrieve the correct neighbors.

\section{Conclusion}

We propose TMM-NN, a new method for nearest neighbour retrieval based
on the deep feature manifold of a deep network. TMM-NN uses a backdoor
based method to create a sharp distortion through a fine-tuning process
around the query point and then seeking the points that are affected
by that. That way we avoid choosing the most optimal feature layer
and measuring distance in a high-dimensional space. Robustness analysis
shows that this method is more robust than other distance-based methods.
Extensive experiments based on the four datasets show the consistency
of our method under noisy query. Further, we used both visual and
LLM based evaluation and demonstrate that TMM-NN retrievals are almost
always more semantically aligned than other methods. Future work will
look into extending the method for different modalities such as text,
audio etc.

\bibliographystyle{unsrtnat}
\phantomsection\addcontentsline{toc}{section}{\refname}\bibliography{aaai2026}

@Article{Bengio2013,
  author    = {Bengio, Yoshua and Courville, Aaron and Vincent, Pascal},
  journal   = {IEEE transactions on pattern analysis and machine intelligence},
  title     = {Representation learning: A review and new perspectives},
  year      = {2013},
  number    = {8},
  pages     = {1798--1828},
  volume    = {35},
  publisher = {IEEE},
}

@InProceedings{Bilgin2021,
  author    = {Bilgin, Zeki and Gunestas, Murat},
  booktitle = {ICAART (2)},
  title     = {Explaining Inaccurate Predictions of Models through k-Nearest Neighbors.},
  year      = {2021},
  pages     = {228--236},
}

@Article{Cover1967,
  author    = {Cover, Thomas and Hart, Peter},
  journal   = {IEEE transactions on information theory},
  title     = {Nearest neighbor pattern classification},
  year      = {1967},
  number    = {1},
  pages     = {21--27},
  volume    = {13},
  publisher = {IEEE},
}

@Article{Gu2019,
  author    = {Gu, Tianyu and Liu, Kang and Dolan-Gavitt, Brendan and Garg, Siddharth},
  journal   = {IEEE Access},
  title     = {Badnets: Evaluating backdooring attacks on deep neural networks},
  year      = {2019},
  pages     = {47230--47244},
  volume    = {7},
  publisher = {IEEE},
}

@Article{Jeyakumar2020,
  author  = {Jeyakumar, Jeya Vikranth and Noor, Joseph and Cheng, Yu-Hsi and Garcia, Luis and Srivastava, Mani},
  journal = {Advances in neural information processing systems},
  title   = {How can i explain this to you? an empirical study of deep neural network explanation methods},
  year    = {2020},
  pages   = {4211--4222},
  volume  = {33},
}

@Article{Kirkpatrick2017,
  author    = {Kirkpatrick, James and Pascanu, Razvan and Rabinowitz, Neil and Veness, Joel and Desjardins, Guillaume and Rusu, Andrei A and Milan, Kieran and Quan, John and Ramalho, Tiago and Grabska-Barwinska, Agnieszka and others},
  journal   = {Proceedings of the national academy of sciences},
  title     = {Overcoming catastrophic forgetting in neural networks},
  year      = {2017},
  number    = {13},
  pages     = {3521--3526},
  volume    = {114},
  publisher = {National Acad Sciences},
}

@Article{Krizhevsky2009,
  author    = {Krizhevsky, Alex and Hinton, Geoffrey and others},
  title     = {Learning multiple layers of features from tiny images},
  year      = {2009},
  publisher = {Toronto, ON, Canada},
}

@Article{LeCun1998,
  author    = {LeCun, Yann and Bottou, L{\'e}on and Bengio, Yoshua and Haffner, Patrick},
  journal   = {Proceedings of the IEEE},
  title     = {Gradient-based learning applied to document recognition},
  year      = {1998},
  number    = {11},
  pages     = {2278--2324},
  volume    = {86},
  publisher = {Ieee},
}

@Article{Netzer2011,
  author = {Netzer, Yuval and Wang, Tao and Coates, Adam and Bissacco, Alessandro and Wu, Bo and Ng, Andrew Y},
  title  = {Reading digits in natural images with unsupervised feature learning},
  year   = {2011},
}

@Article{Papernot1803,
  author  = {Papernot, Nicolas and McDaniel, Patrick},
  journal = {arXiv preprint arXiv:1803.04765},
  title   = {Deep k-nearest neighbors: towards confident, interpretable and robust deep learning (2018)},
  year    = {1803},
}

@Article{Rajani2020,
  author  = {Rajani, Nazneen Fatema and Krause, Ben and Yin, Wengpeng and Niu, Tong and Socher, Richard and Xiong, Caiming},
  journal = {arXiv preprint arXiv:2010.09030},
  title   = {Explaining and improving model behavior with k nearest neighbor representations},
  year    = {2020},
}

@InProceedings{Schroff2015,
  author    = {Schroff, Florian and Kalenichenko, Dmitry and Philbin, James},
  booktitle = {Proceedings of the IEEE conference on computer vision and pattern recognition},
  title     = {Facenet: A unified embedding for face recognition and clustering},
  year      = {2015},
  pages     = {815--823},
}

@Book{Smola2000,
  author    = {Smola, Alexander J},
  publisher = {MIT press},
  title     = {Advances in large margin classifiers},
  year      = {2000},
}

@InProceedings{Stallkamp2011,
  author       = {Stallkamp, Johannes and Schlipsing, Marc and Salmen, Jan and Igel, Christian},
  booktitle    = {The 2011 international joint conference on neural networks},
  title        = {The German traffic sign recognition benchmark: a multi-class classification competition},
  year         = {2011},
  organization = {IEEE},
  pages        = {1453--1460},
}

@InProceedings{Zhang2018,
  author    = {Zhang, Richard and Isola, Phillip and Efros, Alexei A and Shechtman, Eli and Wang, Oliver},
  booktitle = {Proceedings of the IEEE conference on computer vision and pattern recognition},
  title     = {The unreasonable effectiveness of deep features as a perceptual metric},
  year      = {2018},
  pages     = {586--595},
}

@InProceedings{ghosh2025targeted,
  author       = {Ghosh, Banibrata and Harikumar, Haripriya and Venkatesh, Svetha and Rana, Santu},
  booktitle    = {2025 IEEE Conference on Secure and Trustworthy Machine Learning (SaTML)},
  title        = {Targeted Manifold Manipulation Against Adversarial Attacks},
  year         = {2025},
  organization = {IEEE},
  pages        = {427--438},
}

\end{document}